\documentclass[letterpaper]{article} 
\usepackage{aaai25}  
\usepackage{times}  
\usepackage{helvet}  
\usepackage{courier}  
\usepackage[hyphens]{url}  
\usepackage{graphicx} 
\usepackage{natbib}  
\usepackage{caption} 
\usepackage{algorithm}
\usepackage{algorithmic}

\usepackage{amsmath}
\usepackage{amsfonts}
\usepackage{parskip}

\usepackage{newfloat}
\usepackage{listings}
\DeclareCaptionStyle{ruled}{labelfont=normalfont,labelsep=colon,strut=off} 
\floatstyle{ruled}
\newfloat{listing}{tb}{lst}{}
\floatname{listing}{Listing}
\newcount\Comments  
\usepackage{color}
\definecolor{darkgreen}{rgb}{0,0.5,0}
\definecolor{purple}{rgb}{1,0,1}
\newcommand{\kibitz}[2]{\ifnum\Comments=1\textcolor{#1}{#2}\fi}

\DeclareMathOperator*{\argmax}{arg\,max}

\title{Performance Comparisons of Reinforcement Learning Algorithms for \\ Sequential Experimental Design}

\author {
    Yasir Zubayr Barlas\textsuperscript{\rm 1\thanks{Work carried out at City St George's, University of London.}},
    Kizito Salako\textsuperscript{\rm 2}
}
\affiliations {
    \textsuperscript{\rm 1}The University of Manchester\\
    \textsuperscript{\rm 2}City St George's, University of London\\
    yasir.barlas@manchester.ac.uk, k.o.salako@citystgeorges.ac.uk
}

\begin{document}

\maketitle

\begin{abstract}
Recent developments in sequential experimental design look to construct a policy that can efficiently navigate the design space, in a way that maximises the expected information gain. Whilst there is work on achieving tractable policies for experimental design problems, there is significantly less work on obtaining policies that are able to generalise well -- i.e. able to give good performance despite a change in the underlying statistical properties of the experiments. Conducting experiments sequentially has recently brought about the use of reinforcement learning, where an agent is trained to navigate the design space to select the most informative designs for experimentation. However, there is still a lack of understanding about the benefits and drawbacks of using certain reinforcement learning algorithms to train these agents. In our work, we investigate several reinforcement learning algorithms and their efficacy in producing agents that take maximally informative design decisions in sequential experimental design scenarios. We find that agent performance is impacted depending on the algorithm used for training, and that particular algorithms, using dropout or ensemble approaches, empirically showcase attractive generalisation properties.
\end{abstract}

%

\section{Introduction}

In experimental design, we often need to conduct experiments in the most efficient way possible, minimising both the time and costs involved \citep{atkinson1992optimum, ryanboed, rainforth2024modern, Huan_Jagalur_Marzouk_2024}. As it is usually impractical to conduct hundreds of experiments, it is essential to extract as much information as possible from a limited number of experiments. The design of each experiment is crucial here; if done optimally, this provides scientists with a sufficient amount of high quality data, versus large amounts of low quality data \citep{foster-phd}. This form of experimental design is coined \textit{optimal} experimental design, in that one performs experiments optimally in an effort to maximise the information gained from the experiments, while minimising the effort/cost in performing the experiments.

Given that one ought to conduct a sequence of experiments optimally, Bayesian methods \citep{gelman2013bayesian} provide a natural framework for incorporating prior knowledge -- i.e. both domain expertise and knowledge gained from past experimentation -- into future experimental design choices. Operationally, the prior knowledge justifies \emph{prior beliefs} about model parameters of interest $\boldsymbol{\theta}$ -- these parameters, together with a design choice $\xi$ of the next experiment to carry out, define a statistical model that characterises the uncertainty in the outcome $y$ from using design $\xi$. So that, based on outcomes gathered from past experimentation, these prior beliefs are updated into \emph{posterior beliefs} -- as a consequence, some designs become good choices for the next experiment, because they lead to more informative experimental outcomes, while other designs become less desirable choices \citep{rainforth2024modern}. Prior and posterior beliefs are formally captured as prior and posterior probability distributions of $\boldsymbol{\theta}$. When an emphasis is placed on using Bayesian methods to make optimal design choices, this category of experimental design is known as \emph{Bayesian (optimal) experimental design} (BOED) \citep{lindley, rainforth2024modern}. It is typical in BOED to select designs $\xi \in \Xi$ that maximise the expected information gain (EIG) \citep{lindley}, where $\Xi$ is the design space we can select designs from. The EIG from experiment $\xi$ is the expected reduction in Shannon entropy \citep{shannon} when prior beliefs are updated to posterior beliefs, otherwise the mutual information between $y$ and $\boldsymbol{\theta}$ given the design $\xi$, 
\begin{equation}
\label{eq:1}
\text{EIG}(\xi) = \mathbb{E}_{p(\boldsymbol{\theta})p(y \mid \boldsymbol{\theta}, \xi)}[\log p(\boldsymbol{\theta} \mid y, \xi) - \log p(\boldsymbol{\theta})],
\end{equation} where $p(\boldsymbol{\theta})$ is a prior distribution, $p(y \mid \boldsymbol{\theta}, \xi)$ is the likelihood function of the underlying statistical model, and $p(\boldsymbol{\theta} \mid y, \xi)$ is the posterior distribution. Here, $\boldsymbol \theta$ is assumed statistically independent of the design choice $\xi$.

The goal in BOED is to determine an optimal design-choice strategy -- adapting how the next experiment is chosen, given the choices and outcomes of previous experiments -- that ensures the sequence of designs $\xi_{1}, \ldots, \xi_{T}$ chosen by the end of the experimentation (i.e. by the $T$-th experiment) maximises the EIG from these experiments. That is, given the sequential nature of \emph{adaptive} experimental design problems, one can seek an optimal policy to select the best designs during experimentation \citep{huan2016sequentialbayesianoptimalexperimental, foster2021deep, blau2022optimizing}. 
This policy is a mapping, from past design choices and the data collected using these designs, to the next design selected for experimentation.

In recent years, \emph{reinforcement learning} (RL) \citep{franccois2018introduction, Sutton1998} has emerged as a source of algorithms for obtaining approximately optimal policies \citep{blau2022optimizing, lim2022policybasedbayesianexperimentaldesign, shen2023bayesian}. One approach is to train an RL agent\footnote{An agent learns an optimal policy over time by observing rewards received when actions are taken in a changing environment, re-evaluating how good (in terms of expected future rewards) it is to take each action in each state of the environment, and choosing to take more of those actions that (based on the current evaluation of goodness) maximise the expected total reward received (e.g. EIG).} 
to learn the policy via offline training (using simulations of the experimental design problem), then to deploy the agent online for the real experiments to be performed. 
\citet{blau2022optimizing} demonstrates RL's competitive performance, compared with alternative approaches to BOED that amortise the cost of experimentation \citep{foster2021deep, ivanova2021implicit}. 

RL algorithms possess desirable properties, including the flexibility to: 1) tradeoff exploration of the design space and exploitation of learnt design choice patterns; 2) control the experimentation horizon over which optimal behaviour is sought; and 3) handle non-differentiable likelihood functions and discrete design spaces \citep{Sutton1998, blau2022optimizing}. However, RL algorithms can suffer from an inability to generalise well outside of the training distribution \citep{kirk2023survey}.

Whilst there is work on building tractable frameworks to solve experimental design problems \citep{rainforth2024modern}, there is significantly less work aimed at dealing with model misspecification \citep{WALKER20131621} and distributional shift \citep{wiles2022a}. In practice, the statistical model -- that generates the experimental outcomes our agent is trained with -- may differ greatly from that encountered at deployment time\footnote{This is the setting after training, where the trained agent is required to select maximally informative designs in actual experiments. This is also referred to as evaluation time or test time.}. The generalisability of policies, and the robustness of RL algorithms for learning policies that generalise well, is important for the viability of BOED. However, to the best of our knowledge, there has not been significant work studying the impact statistical model changes have on the performance of RL agents used in BOED applications. In this work, we investigate the generalisation capabilities of several agents trained using different RL algorithms that extend the soft actor-critic (SAC) algorithm \citep{haarnoja2019softactorcriticalgorithmsapplications}. Here, for each algorithm considered, we pose the problem of how well the learnt policies generalise in 2 example BOED applications. We utilise the RL formulation by \citet{blau2022optimizing}, and note the following contributions:

\begin{itemize}
    \item A statistically sound assessment of the impact of using certain RL algorithms to solve BOED problems;
    \item An analysis of the generalisation capabilities of RL agents on BOED problems with statistical models that are related to, but differ from, the models the agents were trained on;
    \item The average time required to train agents under certain RL algorithms;
    \item An improved, but costly, combination of the SUNRISE \citep{lee2021sunrise} and DroQ \citep{hiraoka2022dropout} algorithms, that performs well across two experimental design problems.
\end{itemize}

The rest of this paper is structured as follows. Section \ref{sec_relatedwork} covers previous work in the area of policy-based BOED, how our work relates to model misspecification and distribution shift, and recent policy-based approaches for improving generalisability. Section \ref{sec_probform} explains the setup of BOED as a sequential decision-making problem, which can be solved using RL. Section \ref{sec_algorithms} describes the RL algorithms we investigate in our work, and Section \ref{sec_results} provides the results of using these algorithms on two experimental design problems. We conclude with a discussion in Section \ref{sec_conclusions}.

\section{Related Work}
\label{sec_relatedwork}

Early work in policy-based BOED uses approximate dynamic programming for selecting designs, requiring explicit posteriors to be calculated \citep{huan2016sequentialbayesianoptimalexperimental}.
Subsequent work looks at amortising the costs at deployment time by learning a policy network that rapidly selects designs, both for explicit \citep{foster2021deep, blau2022optimizing} and implicit \citep{ivanova2021implicit, lim2022policybasedbayesianexperimentaldesign} likelihood functions. Due to the expensive computational costs in calculating posteriors, and in turn EIG, a contrastive lower-bound on the EIG is optimised to avoid explicit posterior computations \citep{foster2021deep}. In the RL setting, this lower-bound is the reward function, and is computed incrementally for each experiment performed to avoid the sparse reward problem \citep{Sutton1998, blau2022optimizing}. 

Recent work looks at the use of variational posterior approximations \citep{shen2023variationalsequentialoptimalexperimental, blau2024statisticallyefficientbayesiansequential} to form a lower-bound on the EIG instead. Alternative metrics to EIG have also been investigated in the amortised setting \citep{huang2024amortizedbayesianexperimentaldesign}. The generalisability problem our work seeks to address is connected to model misspecification \citep{sloman2022characterizingrobustnessbayesianadaptive, overstall2022bayesian, catanach2023metrics} and distributional shift \citep{kirschner2020distributionally, zhou2021bayesian}. The statistical model we assume is the same one used to generate data during agent training. Therefore, if our assumed model does not accurately represent the true data-generative process, it is misspecified. As a result of misspecification, a distributional shift occurs when the data distribution during training differs from the one observed at test time. 

\citet{shen2023variationalsequentialoptimalexperimental} present a variational methodology for solving a range of problems within BOED, allowing for the usual targeting of EIG, but also other criteria such as those in handling model discrimination \citep{kleinegesse2021gradientbasedbayesianexperimentaldesign} and nuisance parameters \citep{sloman2024bayesian} -- helping us tackle problems related to generalisability. 
These criteria can bring us closer to generalisable agents, but they require more complicated methods to reduce the problem to one that can be solved with RL sequentially.

\citet{ivanova2024stepdad} propose a method for policy refinement at deployment time after offline training is performed, extending the approaches by \citet{foster2021deep} and \citet{ivanova2021implicit} to allow for better generalisability. This approach is applicable when extra computational resources are accessible at deployment time, but this is not always feasible. 
In contrast, we have focused on the question of selecting the best amongst alternative algorithms for offline training to improve generalisability.

\section{Problem Formulation}
\label{sec_probform}

We take the setting of \citet{blau2022optimizing} and use an extension to Markov decision processes (MDPs) \citep{feinberg2002handbook}, known as hidden-parameter MDPs (HiP-MDPs) \citep{doshivelez2013hiddenparametermarkovdecision}, to formulate the sequential BOED problem. We therefore also follow \citet{blau2022optimizing} by optimising a contrastive lower-bound on the EIG, known as the \emph{sequential prior contrastive estimation} (sPCE) \citep{foster2020unified, foster2021deep}. An upper-bound on the EIG also exists, known as the \emph{sequential nested Monte Carlo} (sNMC) \citep{rainforth2018nesting, foster2021deep}.

\subsection{Sequential Experimental Design}

Let $h_{T}=\{(\xi_{1}, y_{1}), \ldots,(\xi_{T}, y_{T})\}$ denote the experimental history up to time $T$, which captures the design $\xi_{t}$ selected and experiment outcome $y_{t}$ of using $\xi_{t}$ for each experiment $t$. We seek to optimise a policy $\pi : \mathcal{H} \rightarrow \Xi$ that maps the history at time $t$ to the next design to select, with $\xi_{t} = \pi(h_{t - 1})$. Each selected design can be thought of as an action in the RL setting. Our setup is in discrete-time.

According to \citet{foster2021deep}, the EIG under a policy $\pi$, and over a sequence of $T$ experiments, is given by
\begin{equation}
\label{eq:2}
\text{EIG}_{T}(\pi) = \mathbb{E}_{p(\boldsymbol{\theta})p(h_{T} \mid \boldsymbol{\theta}, \pi)}\left[ \log\left(\frac{p(h_{T} \mid \boldsymbol{\theta}, \pi)}{p(h_{T} \mid \pi)} \right)\right],
\end{equation} where $p(h_{T} \mid \boldsymbol{\theta}, \pi) = \prod_{t = 1}^{T} p(y_{t} \mid \boldsymbol{\theta}, \xi_{t})$ is the likelihood of the history and $p(h_{T} \mid \pi) = \mathbb{E}_{p(\boldsymbol{\theta})}[p(h_{T} \mid \boldsymbol{\theta}, \pi)]$ is the marginal likelihood of the history. 

Notice how $\text{EIG}_{T}(\pi)$ does not require any (expensive) posterior computations \citep{foster2021deep}, in contrast to the approach by \citet{blau2024statisticallyefficientbayesiansequential}, who reformulate \eqref{eq:2} to instead require the posterior $p(\boldsymbol{\theta} \mid h_{T}, \pi)$ to be computed. 
One issue still remains: the denominator in \eqref{eq:2}, i.e. $p(h_{T} \mid \pi)$, is typically intractable, and it changes with each sample of $\boldsymbol{\theta}$ and $h_{T}$ from the outermost expectation in \eqref{eq:2}. 

\citet{foster2021deep} propose the sPCE and sNMC bounds to bound EIG, using approximations of $p(h_{T} \mid \pi)$ as follows. Given a sample, say $\boldsymbol{\theta}_{0}$ and experimental history $h_{T}$ from $p(\boldsymbol{\theta}, h_{T} \mid \pi)$, 
we draw $L$ independent contrastive samples $\boldsymbol{\theta}_{1:L}$ from the prior $p(\boldsymbol{\theta})$. We estimate $p(h_{T} \mid \pi)$ by first computing the likelihood of the history under each of our contrastive samples $\boldsymbol{\theta}_{1:L}$, then computing $\frac{1}{L + 1} \sum_{\ell = 0}^{L} p(h_{T} \mid \boldsymbol{\theta}_{\ell}, \pi)$ for sPCE, or $\frac{1}{L} \sum_{\ell = 1}^{L} p(h_{T} \mid \boldsymbol{\theta}_{\ell}, \pi)$ for sNMC. 
So, to obtain the sPCE and sNMC bounds respectively, we either include $\boldsymbol{\theta}_{0}$ in our estimate, or we exclude it:
\begin{align}
\text{sPCE}(\pi, L, T) & = \mathbb{E}_{p(\boldsymbol{\theta}_{0:L})p(h_{T} \mid \boldsymbol{\theta}_{0}, \pi)} \left[g(\boldsymbol{\theta}_{0:L}, h_{T})\right], \label{eq:3}\\
\text{sNMC}(\pi, L, T) & = \mathbb{E}_{p(\boldsymbol{\theta}_{0:L})p(h_{T} \mid \boldsymbol{\theta}_{0}, \pi)} \left[f(\boldsymbol{\theta}_{0:L}, h_{T})\right],
\end{align}
where $g(\boldsymbol{\theta}_{0:L}, h_{T})$ and $f(\boldsymbol{\theta}_{0:L}, h_{T})$ are defined by
\begin{align}
g(\boldsymbol{\theta}_{0:L}, h_{T}) & = \log \left[\frac{p(h_{T} \mid \boldsymbol{\theta}_{0}, \pi)}{\frac{1}{L + 1} \sum_{\ell = 0}^{L} p(h_{T} \mid \boldsymbol{\theta}_{\ell}, \pi)}\right], \label{eq:5}\\
f(\boldsymbol{\theta}_{0:L}, h_{T}) & = \log \left[\frac{p(h_{T} \mid \boldsymbol{\theta}_{0}, \pi)}{\frac{1}{L} \sum_{\ell = 1}^{L} p(h_{T} \mid \boldsymbol{\theta}_{\ell}, \pi)}\right].
\end{align}
$g$ is bounded by $\log (L + 1)$, whilst $f$ might be unbounded \citep{foster2021deep}. Since sPCE is bounded and more numerically stable than sNMC, \citet{foster2021deep} and \citet{blau2022optimizing} train their policies using sPCE. As $L$ increases, the bounds converge to $\text{EIG}_T$, under mild conditions and at higher computational costs \citep{foster2021deep}. 

\subsection{Hidden-Parameter Markov Decision Process}

The HiP-MDP \citep{doshivelez2013hiddenparametermarkovdecision} extends the MDP \citep{feinberg2002handbook} by enabling rewards and transition functions to be parameterised by the model parameter $\boldsymbol{\theta}$ of the underlying statistical model for the experiments, allowing the sequential BOED setting to be cast as a ``parameterised'' MDP problem. For this, because the true parameter value is unknown, one defines a prior distribution over the model parameter space $\boldsymbol{\Theta}$. At the beginning of each training episode when solving the MDP, parameter values $\boldsymbol{\theta}_{0:L}$ are sampled from $\boldsymbol{\Theta}$ according to this prior -- these are the only parameter values used until the final experiment at time $T$. New parameter samples are drawn for each episode \citep{blau2022optimizing}.

To formalise the HiP-MDP setup, along the lines of \citet{blau2022optimizing}, the following function is required for the state space, transition dynamics, and reward function. 
Let the vector of history likelihoods $C_{t}$ be defined by
\begin{align*}
C_{t} & = \left[\prod_{k = 1}^{t} p(y_{k} \mid \boldsymbol{\theta}_{\ell}, \xi_{k}) \right]^{L}_{\ell = 0}.
\end{align*}

A sequential BOED problem is formalised within the HiP-MDP framework as a tuple, $\langle \mathcal{S}, \mathcal{A}, \Theta, \mathcal{T}, \mathcal{R}, s_0, \gamma, P_{\Theta}\rangle$, with each element of the tuple being derived from the experimental design problem:

\begin{itemize}
    \item $\mathcal{S}$ is the state space, which is the set of all tuples $s_{t} \in \mathcal{S}$, where $s_{t} = (h_{t}, C_{t}, y_{t})$ $\forall t \in [0, T]$ and we define $s_{0} = (\emptyset, \boldsymbol{1}, \emptyset)$ since there is no history or experimental outcome initially;
    \item $\mathcal{A}$ is the action space, which is the set of all possible designs $\Xi$, and $a_{t} \in \mathcal{A}$, where $a_{t - 1} = \xi_{t} = \pi(h_{t - 1})\,$ $\forall t \in [1, T]$;
    \item $\Theta$ is the model parameter space, containing all of the possible parameter values for the dynamics of the model;
    \item The prior distribution, $P_{\boldsymbol{\Theta}} = p(\boldsymbol{\theta})$, of the model parameters. 
    For our sequential BOED problems, $\boldsymbol{\theta}_{0:L} \overset{\text{i.i.d.}}{\sim} p(\boldsymbol{\theta})$;
    \item $\mathcal{T}$ is the transition dynamics for the model -- a mapping from the current state to a new state when an action is taken in the current state. For $\xi_t=\pi(h_{t-1})$ and state $s_{t-1}$, the new state $s_t$ is determined by
    \begin{align*}
    y_{t} & \sim p(y_{t} \mid \boldsymbol{\theta}_{0}, \xi_{t}), \\ 
    C_{t} & = C_{t - 1} \odot [p(y_{t} \mid \boldsymbol{\theta}_{\ell}, \xi_{t})]^{L}_{\ell = 0}, \\
    h_{t} & = h_{t-1} \cup \{(\xi_{t},y_{t})\},
    \end{align*}
    with Hadamard product $\odot$. The Markov property holds: transition functions determine state $s_{t}$ from $s_{t-1}$;
    \item $\mathcal{R}$ is the reward function, which provides a reward for transitioning to another state by taking a certain action:
    \begin{align*}
    r_{t} & = \mathcal{R}(s_{t - 1}, a_{t - 1}, s_{t}; \boldsymbol{\theta}_{0:L}) \\
    & = \log p(y_{t} \mid \boldsymbol{\theta}_{0}, \xi_{t}) - \log(C_{t} \cdot \boldsymbol{1}) + \log(C_{t - 1} \cdot \boldsymbol{1})\,,
    \end{align*}
    where $C_{t} \cdot \boldsymbol{1}$ is a dot product with a vector of ``$1$''s;
    \item $\gamma \in [0, 1]$ is the discount factor, determining the importance of future rewards over immediate ones. 
\end{itemize}

Using the above, the state-action value function $Q^{\pi}(s_{t}, a_{t})$, resulting from taking actions according to policy $\pi$, is
\begin{equation*}
Q^{\pi}(s_{t}, a_{t}) = \mathbb{E}_{\mathcal{T}, \pi, P_{\boldsymbol{\Theta}}}\left[G_{t} \mid s_{t}, a_{t}\right],
\end{equation*}
where $G_{t} = \sum_{u = t}^{T-1}\gamma^{u-t}r_{u+1}$ is the total discounted reward from timestep $t\in [0,T-1]$. The optimal policy, $\pi^\ast$, and the state-action value function obtained by following this policy, $Q^\ast$, are the unique pair such that the policy chooses actions which maximise the value function in each state; that is, 
$$\pi^\ast(h_{t}) = \argmax\limits_{a\in\mathcal{A}} Q^\ast(s_{t}, a).$$
In sequential BOED terms, we have
$$\max\limits_{\pi}\text{sPCE}(\pi,L,T) =\text{sPCE}(\pi^\ast,L,T) = Q^\ast(s_0,\pi^\ast(\emptyset)).$$

The optimal policy maximises the sPCE, since the definition of the reward function and $G_t$ (assuming $\gamma=1$) imply the expected total discounted rewards (from time $t=0$) equals the sPCE \citep{blau2022optimizing}. This definition of $\mathcal R$ mitigates the sparse reward problem in RL \citep{Sutton1998}.

To find the optimal policy we utilise deep RL algorithms\footnote[3]{In the experiments we conduct using these algorithms, we assume $\gamma$ values close to $1$ (rather than $\gamma=1$) for computational stability; this follows \citet{blau2022optimizing}.} \citep{franccois2018introduction}. The policy is implemented as a deep neural network that relies on a summary of $h_t$ in the form of an encoder network. Although the definition of the HiP-MDP satisfies the Markov property, it does so by explicitly including $h_t$ as part of $s_t$; the use of an encoder network weakens this by not explicitly requiring $h_t$ to be part of the state. Instead, the encoder network is recursively defined, so that a summary representation of $h_t$ at time $t$ only depends on the summary representation at time $t-1$ and $(\xi_t,y_t)$ \citep{blau2022optimizing}. A detailed description of the policy network, as well as more details on the HiP-MDP implementation, are given in Appendix \ref{App_A}.  

\section{Algorithms}
\label{sec_algorithms}

The (model-free) RL algorithms we employ are REDQ \citep{chenrandomized}, DroQ \citep{hiraoka2022dropout}, SBR \citep{doro2022sampleefficient}, and SUNRISE \citep{lee2021sunrise}. These all extend the SAC algorithm \citep{haarnoja2019softactorcriticalgorithmsapplications} by proposing improvements to the training procedure. Appendix B contains the full explanation and pseudocode of each algorithm.

\textbf{REDQ}: \emph{Randomised ensemble double Q-learning} (REDQ) \citep{chenrandomized} is a model-free algorithm that looks to employ a higher update-to-data (UTD) ratio and the use of a large ensemble of Q-functions. The idea here is that model-based methods such as model-based policy optimisation \citep{janner2019trust} use a higher UTD ratio, which is the number of updates taken by the agent compared to the number of actual interactions with the environment. This higher UTD ratio allows for greater sample efficiency. REDQ is the algorithm used in the approach by \citet{blau2022optimizing}.

\begin{table*}[ht!]
    \centering
    \begin{tabular}{cccccccc}
         & \multicolumn{5}{c}{sPCE} & Time \\
        \hline
        & $K = 1$ & $K = 2$ & $K = 3$ & $K = 4$ & $K = 5$ & $K = 2$ \\
        \hline
        REDQ & $6.279 \pm 0.013$ & $11.689 \pm 0.012$ & $11.881 \pm 0.013$ & $11.507 \pm 0.015$ & $11.095 \pm 0.017$ & \textbf{13.23h} \\
        SBR & $6.165 \pm 0.012$ & $11.362 \pm 0.013$ & $11.788 \pm 0.014$ & $11.518 \pm 0.016$ & $11.257 \pm 0.017$ & 13.52h \\
        DroQ & $6.368 \pm 0.013$ & $11.680 \pm 0.013$ & $11.902 \pm 0.013$ & $11.480 \pm 0.015$ & $11.090 \pm 0.017$ & 19.05h \\
        SUNRISE & $6.340 \pm 0.013$ & \textbf{11.837 $\pm$ 0.012} & $12.133 \pm 0.013$ & \textbf{11.846 $\pm$ 0.014} & $11.445 \pm 0.016$ & 21.44h \\
        SUNRISE-DroQ & \textbf{6.433 $\pm$ 0.012} & $11.770 \pm 0.012$ & \textbf{12.143 $\pm$ 0.013} & $11.831 \pm 0.014$ & \textbf{11.453 $\pm$ 0.016} & 29.77h \\
        Random & $5.057 \pm 0.010$ & $7.443 \pm 0.016$ & $8.820 \pm 0.018$ & $9.730 \pm 0.019$ & $10.346 \pm 0.018$ & - \\
    \end{tabular}
    \caption{sPCE at $T = 30$ computed using $L = 1e6$, and average agent training time for the location finding experiment. Means and standard errors are from 20000 rollouts, spread evenly across 10 agents trained under unique random seeds. The average training time is across these agents. $T$ and $L$ differ for Table~\ref{overallspce2}, and the statistics are calculated in the same way.}
\label{overallspce1}
\end{table*}

\textbf{DroQ}: \emph{Dropout Q-functions for doubly efficient RL} (DroQ) \citep{hiraoka2022dropout} is an algorithm that seeks to improve on the computational costs of REDQ by introducing dropout regularisation \citep{hinton2012improving} and layer normalisation \citep{ba2016layer} in the Q-functions. REDQ uses a large ensemble of Q-functions to reduce estimation bias, which is crucial for high sample efficiency. But despite REDQ's advantages, it is computationally intensive due to needing large numbers of Q-functions, and it is expensive to update these Q-functions. \citet{hiraoka2022dropout} through DroQ find empirically competitive performance with a much smaller ensemble compared to REDQ.

\textbf{SBR}: Sample-efficient RL by breaking the replay ratio barrier, or \emph{scaled-by-resetting} (SBR) \citep{doro2022sampleefficient}, explores how increasing the replay ratio -- the number of times an agent's parameters are updated per environment interaction -- can drastically improve the sample efficiency of RL algorithms. Essentially, the parameters of the neural networks for the agent are periodically reset fully during training. This resetting mechanism allows the agents to better handle high replay ratios, enabling them to undergo more updates without degrading their learning and generalisation capabilities.

\textbf{SUNRISE}: The \emph{simple unified framework for RL using ensembles} (SUNRISE) algorithm \citep{lee2021sunrise} addresses common challenges like instability in Q-learning and improving exploration. It achieves this by integrating ensemble-based weighted Bellman backups, which re-weight target Q-values based on uncertainty estimates from a Q-ensemble, and an upper-confidence bound (UCB) based exploration strategy that selects actions with the highest UCB to encourage efficient exploration. SUNRISE uses an ensemble of agents and their respective Q-functions, instead of solely an ensemble of Q-functions like REDQ and DroQ.

\textbf{SUNRISE-DroQ}: After reviewing how the trained agents from these algorithms performed when deployed, we decided to combine SUNRISE with the use of dropout Q-functions in DroQ. Potentially, this takes advantage of the ensemble-based SUNRISE with UCB exploration \citep{lee2021sunrise}, and the regularisation benefits of dropout \citep{hiraoka2022dropout}.

\section{Experiments}
\label{sec_results}

\citet{blau2022optimizing} show empirically that their approach performs better than the non-RL amortised approach by \citet{foster2021deep}, and offers state-of-the-art performance on other baselines \citep{blau2024statisticallyefficientbayesiansequential}. For this reason, we only compare the performance of RL algorithms and not the other approaches. What follows is a summary of how these RL algorithms performed in two experimental design problems. Our training regimes are explained in Appendix C.

We note that our experimental design problems do not have terminal states, and so fixing a certain budget to conduct experiments is standard practice. A limited budget is often required when performing experiments in real-time, and this also forces the agent to make informed design choices.

\subsection{Location Finding}

The location finding experiment was explored by \citet{foster2021deep} and \citet{blau2022optimizing}. It is known there are $2$ objects placed in a 2-dimensional space, and the aim is to determine the unknown locations of these objects; denote the locations $\boldsymbol{\theta} = \{\boldsymbol{\beta}_{i}\}_{i = 1}^{K}$, for $K=2$. Each object emits a signal which obeys an inverse-square law. We need to select designs $\xi$, which are coordinates in the space, in an effort to find the locations of the objects based on the observed signals. The signal strength of a single object increases as we select $\xi$ closer to this object, and it decays as we choose $\xi$ farther away. As we have multiple objects, we instead observe the total signal intensity, which is a superposition of the signals emitted by the individual objects. Hyperparameter details can be found in Appendix C, and experiment details can be found in Appendix D.1.

Table~\ref{overallspce1} displays the sPCE values across a number of experimental setups with varying numbers of $K$ objects to identify the locations of. See Appendix C for sNMC results. The agents are trained to maximise sPCE in $T = 30$ experiments for $K = 2$ objects, and do not encounter the other values for $K$ in Table~\ref{overallspce1} during training. 
This presents a set of challenging scenarios, as the agents attempt to maximally gather information at deployment time based on their knowledge of navigating an environment with only 2 objects. To perform well when deployed, they need to have learnt a generalisable policy to sensibly select designs for experimentation. Such problems can arise in practical scenarios where a zero-shot RL agent \citep{kirk2023survey} is required to navigate a space with fewer or many more objects -- perhaps due to an incorrect assumption on the number of objects during training. 

Excluding SUNRISE-DroQ, SUNRISE offers the best performance except when $K = 1$, where it falls behind DroQ. SBR outperforms REDQ for $K \in \{4, 5\}$, which may point towards better generalisation in experiments with larger numbers of objects. SUNRISE and DroQ offer the best performances, but they are relatively expensive algorithms. One may argue that the training time is worth the wait due to the performance increases, particularly in the case of SUNRISE.

By combining SUNRISE and DroQ, we find further performance increases for $K = \{1, 3, 5\}$. The combination slightly falls behind SUNRISE for the other values of $K$. One can claim that our combined algorithm offers the greatest performance, at the sacrifice of lower sPCE values for $K = 2$ and $K = 4$. The training time is unfortunately the highest as a result of combining two already expensive algorithms, so one could argue against training the combined algorithm in favour of SUNRISE. Using weighted Bellman backups seems to be advantageous here for SUNRISE and our combination, not least due to the UCB exploration performed as a result of training more than one agent. Dropout regularisation performs best with very small dropout probabilities, meaning that dropping out many neural network nodes during training is disadvantageous here.

\begin{table*}[ht!]
    \centering
    \begin{tabular}{cccccccc}
         & \multicolumn{2}{c}{sPCE} & \multicolumn{2}{c}{sNMC} & Time \\
        \hline
        & $\nu = 0.005$ & $\nu = 0.01$ & $\nu = 0.005$ & $\nu = 0.01$ & $\nu = 0.005$ \\
        \hline
        REDQ & $13.782 \pm 0.021$ & $12.343 \pm 0.022$ & $19.911 \pm 0.155$ & $13.433 \pm 0.046$ & \textbf{8.95h} \\
        SBR & $13.640 \pm 0.022$ & $12.241 \pm 0.023$ & $20.567 \pm 0.179$ & $13.519 \pm 0.052$ & 9.03h \\
        DroQ & \textbf{14.090 $\pm$ 0.020} & \textbf{12.683 $\pm$ 0.022} & \textbf{21.764 $\pm$ 0.201} & \textbf{14.108 $\pm$ 0.058} & 13.18h \\
        SUNRISE & $13.725 \pm 0.022$ & $12.381 \pm 0.023$ & $21.429 \pm 0.199$ & $13.812 \pm 0.058$ & 15.11h \\
        SUNRISE-DroQ & $13.938 \pm 0.021$ & $12.544 \pm 0.022$ & $21.210 \pm 0.165$ & $13.875 \pm 0.049$ & 18.16h \\
        Random & $11.145 \pm 0.035$ & $9.692 \pm 0.030$ & $12.047 \pm 0.056$ & $9.796 \pm 0.032$ & -
    \end{tabular}
    \caption{sPCE and sNMC at $T = 10$ computed using $L = 1e7$, and average agent training time for the CES experiment.}
    \label{overallspce2}
\end{table*}

\begin{figure}[ht!]
    \centering
    \includegraphics[width=80mm]{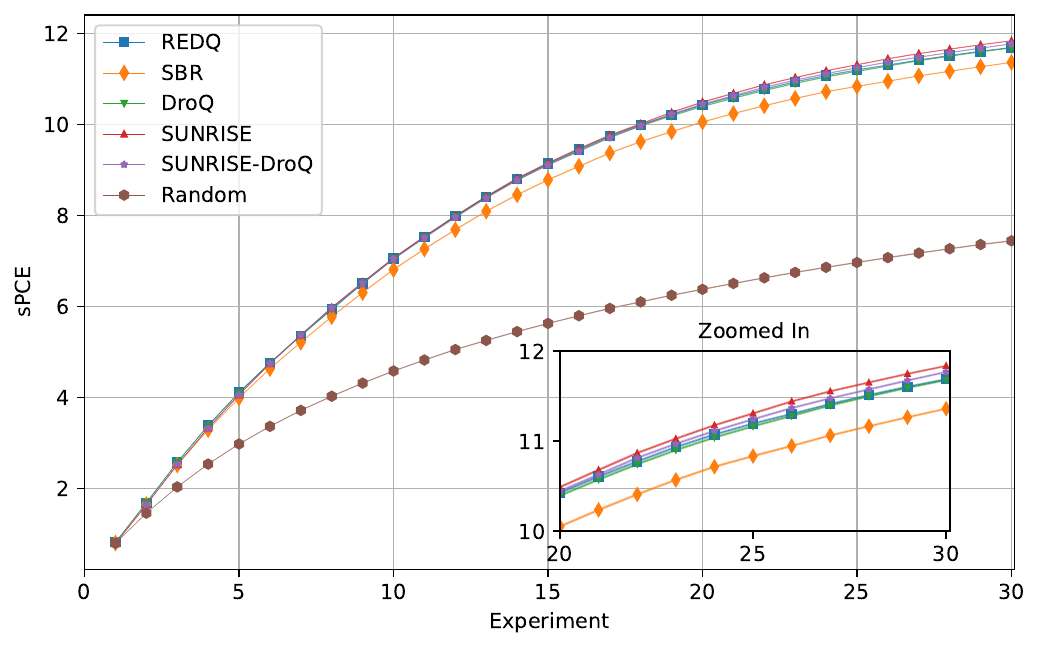}
    \caption{Cumulative sPCE for $K = 2$ objects estimated using $L = 1e6$. Points are mean sPCE values, and the thickness of the shaded regions at each point are standard errors, all from 20000 rollouts, spread evenly across 10 agents trained under unique random seeds. In both Figures~\ref{overallres1} and \ref{overallres2}, interpolations and shaded regions between experiment points on the curves are not meaningful, and are only present to emphasise the shading for standard errors.}
    \label{overallres1}
\end{figure}

\begin{figure}[ht!]
    \centering
    \includegraphics[width=80mm]{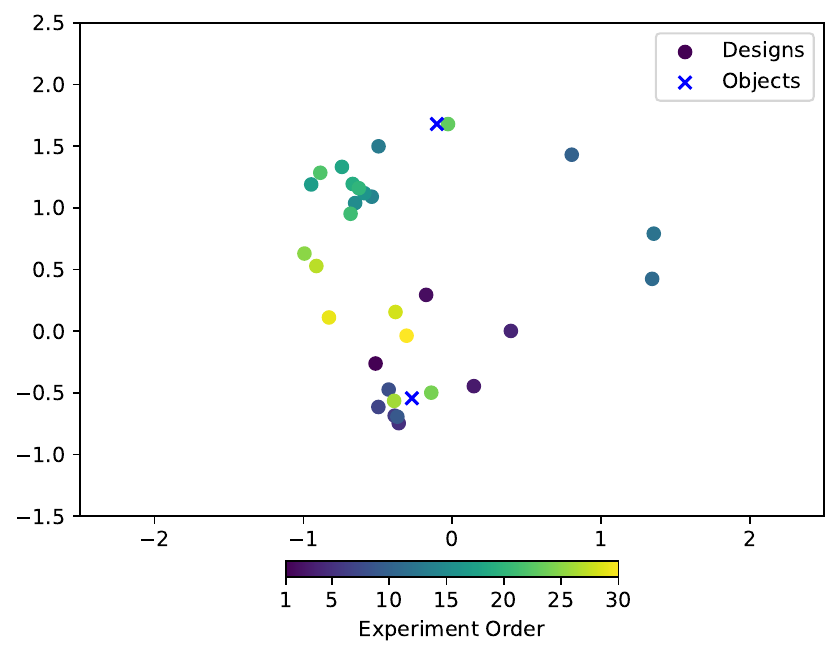}
    \caption{Trained REDQ agent selecting designs at deployment time in the location finding experiment.}
    \label{loc_find_example}
\end{figure}

Observing Figure~\ref{overallres1}, at deployment time for $K = 2$, which is the same setup used during training, the agents all initially start by collecting a similar sPCE value in the first experiment. They follow the same rate, with the first agents to deviate from the pattern being that of an agent randomly choosing designs from the 2nd experiment, and SBR from about the 7th experiment. The random agent ultimately fails to capture the much higher EIG values that the other agents collect. SUNRISE performs best on the same setup used during training, and very well on other numbers of objects.

\subsection{Constant Elasticity of Substitution}

The constant elasticity of substitution (CES) experiment was explored by \citet{foster2019variational}, \citet{foster2020unified} and \citet{blau2022optimizing}. We have two baskets $\boldsymbol{x}, \boldsymbol{x'} \in [0, 100]^{3}$ of goods, and a human indicates their preference of the two baskets on a sliding 0-1 scale. The CES model \citep{arrowchen} with latent variables $\boldsymbol{\theta} = (\rho, \boldsymbol{\alpha}, u)$, which all characterise the human’s utility for the different items, is then used to measure the difference in utility of the baskets. The goal is to design the baskets in a way that allows for inference of the latent variables. The baskets are 3-tuples, meaning that we have $6$ design space dimensions $(\xi = (\boldsymbol{x}, \boldsymbol{x'}))$. Hyperparameter details can be found in Appendix C, and experiment details can be found in Appendix D.2.

Table~\ref{overallspce2} explains the sPCE and sNMC values across two different experimental setups, which vary according to a parameter $\nu$ in the statistical model. Here, the likelihood function used in training differs from that found at deployment time. The training and test distributions are therefore different when $\nu$ is changed, which is a common scenario in the real-world. The agents are trained to maximise sPCE in $T = 10$ experiments for $\nu = 0.005$. To generalise to an unseen statistical model, an agent should still be able to make maximally informative design choices for the two baskets.

\begin{figure}[ht]
    \centering
    \includegraphics[width=80mm]{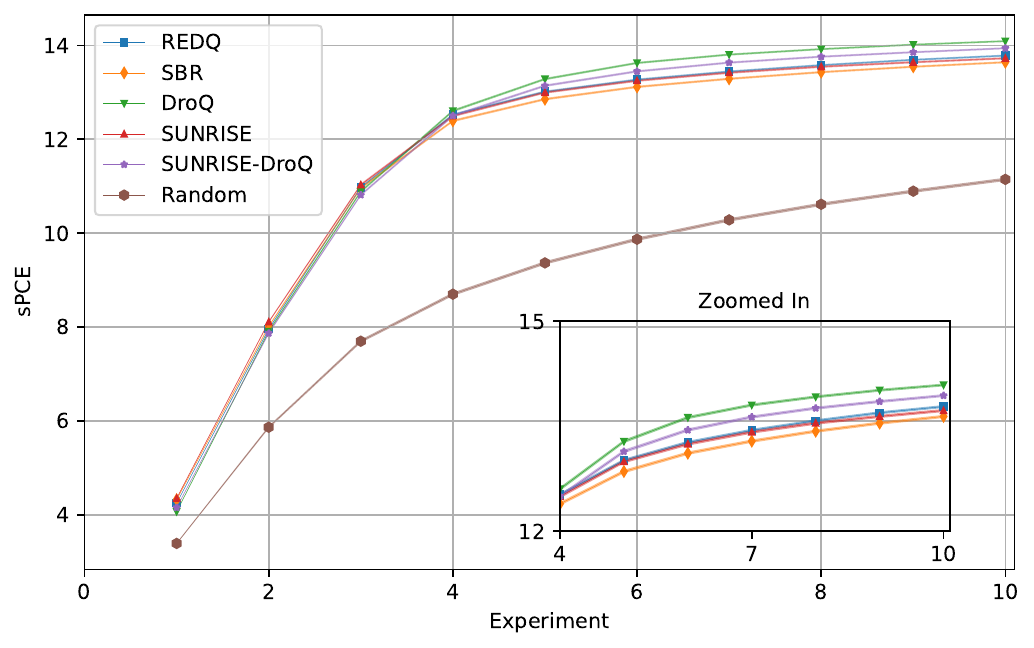}
    \caption{Cumulative sPCE for $\nu = 0.005$ estimated using $L = 1e7$.}
    \label{overallres2}
\end{figure}

DroQ achieves the best performance on both $\nu = 0.005$ and $\nu = 0.01$, noting its ability to generalise and sacrifice short-term gains  
for long-term ones, which we find in Figure~\ref{overallres2}. 
SUNRISE performs very well in the location finding experiment, but here it lags behind DroQ. 
REDQ presents greater sPCE for $\nu = 0.005$ than SUNRISE, and so it is possible that because SUNRISE only uses a single Q-function per agent, it does not tackle overestimation bias \citep{fujimoto2018addressing} in the CES experiment very well. For SUNRISE, there could be better alternatives to randomly switching between multiple agents during experimentation; see Appendix F. The difference between REDQ and DroQ is the use of dropout regularisation for the Q-function neural network, which seems to affect performance greatly here. We find that REDQ offers the cheapest training times by about 4 hours compared to DroQ. We also note that DroQ offers larger standard errors for sNMC, compared to the other algorithms. This is not of great concern here, since the estimated sPCE values from DroQ are 2 orders of magnitude larger than the corresponding standard errors, and the $95\%$ confidence intervals -- based on the standard errors and centred on the DroQ sNMC estimates -- do not overlap with the analogous confidence intervals from the other algorithms. However, when computing DroQ estimates of sNMC in other applications, it would be prudent to monitor for unacceptably large standard errors.

Combining SUNRISE and DroQ does not lead to improved performance over DroQ, but there is an improvement over every other algorithm in terms of sPCE. This is due to using dropout Q-functions in the ensemble. The combination requires much more training time.

\section{Discussion}
\label{sec_conclusions}

We investigated five RL algorithms in an effort to produce agents that perform well when the distribution of experimental observations differs from the distribution that the agents were trained under. We examined REDQ, DroQ, SBR, SUNRISE, and SUNRISE-DroQ, finding the best generalisation from DroQ and SUNRISE, depending on the experimental design problem. REDQ can be viewed as a cheap algorithm for obtaining informative experiments, but it loses out to the greater information gain acquired by more computationally expensive algorithms. A scientist should determine the available computational resources for training, before making a judgement on the algorithm they wish to employ.

Overall, the results suggest the SUNRISE-DroQ combination is a good choice for training agents (on possibly misspecified models) that are expected to achieve high information gain in sequential experiments. The combination does come with added expense determined by ensemble size and other algorithm hyperparameters. 

These results are indicative, as they suggest which of these algorithms give the best performance on these experimental design problems. However, a number of questions remain. How close to optimal are the policies learnt and what factors strongly influence \emph{which} policies are (not) learnt? For example, in the location finding experiment, the different RL algorithms appear to converge on similar policies; policies that tend to initially choose designs close to the centre of the experiments' square-shaped domain. Are such policies close to optimal, and/or are they suggestive of the influence of the radial symmetry in the standard Gaussian prior distribution of the unknown locations? 
The policy might also be influenced by the radial symmetry in the standard Gaussian distribution of the initial placement of objects at unknown locations. 
If, instead, training used uninformative, high-entropy distributions (e.g. uniform distributions), this might result in learnt policies that perform better across a wider range of location finding problems. 
On the other hand, the results might confirm that radial symmetry in initial design choices is necessary for optimal information gain in location finding.

Table~\ref{overallspce1} suggests that the algorithm with the best performing agents (on $K=2$ experiments) is not necessarily producing agents that are the best under distribution drift (on $K\neq 2$ experiments). The results also suggest there are experimental design problems where a validated, highly performant agent can be expected to remain highly performant under distribution drift. Gaining a good understanding of why, and when, this is the case, is important for applying these algorithms in practice. The SUNRISE and (more expensive) SUNRISE-DroQ algorithms exemplify these observations.  

Figures~\ref{overallres1} and \ref{overallres2} show larger jumps in cumulative sPCE from earlier experiments, compared to jumps from latter experiments. Are the policies myopically gathering as much information as possible early on, or are the averages obfuscating more sophisticated design choice strategies? There is some subtle evidence in Figure~\ref{overallres2} that the DroQ agents might be gathering less information than the other RL agents initially, but eventually DroQ starts dominating them later. 

Scientists may be able to afford to conduct additional experiments at deployment time. So, seeing how the RL agents perform over longer horizons would be worth investigating.

For the CES experiment, agents initially appear to maximise sPCE a lot faster than the agents in the location finding experiment (compare Figures~\ref{overallres1} and \ref{overallres2}). 
However, the CES experiment agents appear to converge much more slowly to the sPCE theoretical maximum for the CES experiment, $\log(1e7+1)$, compared to the analogous convergence for the location finding agents to $\log(1e6+1)$. 
This is likely to be an exploration problem, given the small budget. Future work could investigate how certain likelihood function parameters, for both experimental design problems, affect the final sPCE obtained by the agents. 
Another study could look at the performance of these algorithms on higher dimensional scenarios; e.g. 3-dimensional location finding experiments. 
One could also craft utility functions that can better address model misspecification and lead to more robust experimental design \citep{go2022robust, catanach2023metrics}. 

To conclude, our results showcase an important step towards generalisable RL agents for BOED. We extend the work of \citet{blau2022optimizing} by considering alternative RL algorithms, in aid of constructing generalisable policies. 

\section*{Acknowledgements}

YZB acknowledges support by G-Research through the ELLIS Mobility Fund to attend the ELLIS Doctoral Symposium 2024, to present some of the early findings of our work. YZB is supported by a departmental studentship at The University of Manchester. All experiments were run on the Hyperion High-Performance Computer at City St George's, University of London. The authors thank Sabina Sloman for recommending a submission to the `8th Workshop on Generalization in Planning' at AAAI 2025, and for suggesting changes to an initial draft of this paper.

\bibliography{aaai25}

\newpage
\appendix
\onecolumn

\section{\raggedright{Neural Network Architecture and the HiP-MDP}}
\label{App_A}

We explain how the design neural networks are constructed in our experiments and how this relates to our HiP-MDP formulation.

\subsection{Neural Network Architecture}

The optimal policy is given by $\pi^{*} = \argmax\limits_{\pi} \text{sPCE}(\pi, L, T)$, which maximises the sPCE lower-bound. Recall that sPCE is bounded by $\log(L + 1)$, and so the optimal policy should achieve this (as a reward) by the end of experimentation. The approach by \citet{foster2021deep}, which they coin \textit{deep adaptive design} (DAD), seeks to approximate the optimal policy $\pi^{*}$ through a neural network, otherwise known as the \textit{design network} $\pi_{\phi}$. This is then a policy-based approach to sequential BOED, since we now learn and use a policy $\pi$ to select designs, rather than optimising designs at the same time during experimentation. This policy would be learnt offline using the model provided of the Bayesian experimental design problem, and would then be deployed in the online setting for rapid live experiments.

We say that DAD \textit{amortises} the cost of conducting experiments in experimental design, since we are now learning the parameters $\phi$ of a neural network $\pi_{\phi}$, rather than optimising designs directly during training. This approach eliminates the adaptation costs during the live experiment, as the design network can instantly select the next design with a single forward-pass through the neural network.

In DAD, the neural network architecture for $\pi_{\phi}$ is constructed to allow for permutation invariance, enhancing the efficiency of learning by allowing for weight sharing \citep{foster2021deep}. Using the fact that EIG is unchanged under permutation \citep{foster2021deep}, we represent the history $h_{t}$ with a fixed-dimensional representation by pooling representations of the distinct design-observation pairs of the history,$$B_{\psi, t} = \sum_{k = 1}^{t} \text{Encoder}_{\psi} (\xi_{k}, y_{k}),$$ where $\text{Encoder}_{\psi}$ is a neural network \textit{encoder} with parameters $\psi$ to be learnt. This pooled representation remains unchanged if we reorder the labels $1, \ldots, t$.

We then construct our design network to make decisions based on the pooled representation $B_{\psi, t}$ by setting, $$\pi_{\phi}(h_{t}) = \text{Emitter}_{\eta} (B_{\psi, t}),$$ where $\text{Emitter}_{\eta}$ is a learnt \textit{emitter} network. The trainable parameters are $\phi = \{\psi, \eta\}$. By combining simple networks in a way that respects the permutation invariance of the problem, we enable parameter sharing \citep{foster2021deep}. This allows the network $\text{Encoder}_{\psi}$ to be reused for each input pair and for each timestep $t$.

The exact architecture of our design network, including the number of nodes and hidden layers, that we use in our results is provided in Appendix C.1.

\subsection{HiP-MDP}

\citet{blau2022optimizing} explain that each state $s_{t}$ in the state space $\mathcal{S}$ has the form $s_{t} = (B_{\psi, t}, C_{t}, y_{t})$, where $$B_{\psi, t} = \sum_{k = 1}^{t} \text{Encoder}_{\psi}(\xi_{k}, y_{k}), \hspace{1ex} C_{t} = \left[\prod_{k = 1}^{t} p(y_{k} \mid \boldsymbol{\theta}_{\ell}, \xi_{k}) \right]^{L}_{\ell = 0},$$ and $\text{Encoder}_{\psi}$ is our encoder in the design network with parameters $\psi$, all as explained in Appendix A.1.

This summary of the history (or history summary) follows the Markov property because we can decompose this as $$B_{\psi, t} = B_{\psi, t - 1} + \text{Encoder}_{\psi}(\xi_{t}, y_{t}).$$ In other words, we do not need the whole history $h_{t} = (\xi_{1:t}, y_{1:t})$ to compute the next summary $B_{\psi, t + 1}$. It is then sensible to use $B_{\psi, t}$ as an input for the policy, or in other words, construct a pooled representation of the history $h_{t}$, following the Markovian representation above, and pass this through an emitter network (see Appendix A.1) to obtain the next design to select. The initial state becomes $s_{0} = (\boldsymbol{0}, \boldsymbol{1}, \emptyset)$. The transition dynamics for $B_{\psi, t}$ follow the Markovian representation above.

Recall that sPCE requires $L$ contrastive samples, being different realisations of the parameter of interest $\boldsymbol{\theta}$. We can compute the sPCE for experiment $t$ as $r_{t} = \mathcal{R}(s_{t - 1}, a_{t - 1}, s_{t}; \boldsymbol{\theta}_{0:L})$, where $$\mathcal{R}(s_{t - 1}, a_{t - 1}, s_{t}; \boldsymbol{\theta}_{0:L}) = \log p(y_{t} \mid \boldsymbol{\theta}_{0}, \xi_{t}) - \log(C_{t} \cdot \boldsymbol{1}) + \log(C_{t - 1} \cdot \boldsymbol{1}).$$
$C_{t}$ here is a vector of history likelihoods, where each element $c_{t, l}$ is the likelihood of observing the history $h_{t}$ for true parameters $\boldsymbol{\theta}_{l}$. $\boldsymbol{0}$ and $\boldsymbol{1}$ are vectors of zeroes and ones respectively, having the same length as $C_{t}$ or $C_{t + 1}$ where sensible. This reward is generally non-zero for each timestep, and there is no need to save the entire history in memory to compute the reward \citep{blau2022optimizing}. Retaining the history likelihood $C_{t}$ and experimental outcome $y_{t}$ is sufficient.

Using the above, \citet{blau2022optimizing} prove that the expected return of the HiP-MDP is equivalent to sPCE using $\gamma = 1$. The proof is omitted here, but can be found in Appendix A.2 by \citet{blau2022optimizing}.

\section{\raggedright{Algorithms}}

In this appendix, we cover our chosen RL algorithms in more detail, with their respective pseudocode.

\subsection{Soft Actor-Critic}

It is first sensible to cover the foundational algorithm that our explored variants are based on, the SAC algorithm \citep{haarnoja2019softactorcriticalgorithmsapplications}.

SAC is an off-policy RL algorithm designed to optimise a stochastic policy in an environment, balancing exploration and exploitation \citep{haarnoja2019softactorcriticalgorithmsapplications}. SAC aims to maximise both the expected reward and the entropy of the policy. Higher entropy encourages exploration by promoting more stochastic policies, preventing premature convergence to suboptimal policies. This is important in experimental design, as we need to adequately explore the design space to find the appropriate designs that maximise EIG.

SAC follows from maximum-entropy RL \citep{eysenbach2022maximum}, with the following objective $$\pi^* = \argmax\limits_{\pi} \sum_{t} \mathbb{E}_{(s_t, a_t) \sim \rho_\pi} \left[ r_t + \alpha H(\pi(\cdot \mid s_t)) \right],$$ where $\alpha$ is the temperature parameter controlling the importance of the entropy term, and $H$ is the entropy term. A higher $\alpha$ value encourages more exploration by making the policy more random, whilst a lower $\alpha$ value leads to more exploitation by making the policy more deterministic.

\citet{haarnoja2019softactorcriticalgorithmsapplications} derive soft policy iteration, which learns optimal maximum-entropy policies by alternating between policy evaluation and policy improvement in the maximum-entropy framework. This is based on the tabular setting, and so it is expensive to utilise on continuous settings. This brings us to SAC, which uses neural networks to approximate the soft Q-function $Q_{\theta}$ and the policy $\pi_{\phi}$, with parameters $\theta$ and $\phi$ respectively, and optimises them using stochastic gradient descent.

The soft Q-function is updated by minimising the soft Bellman residual, using samples from a replay buffer $D$ and a target soft Q-function with parameters $\theta_{\text{targ}}$, $$J_Q(\theta) = \mathbb{E}_{(s_t, a_t) \sim D} \left[ \frac{1}{2} \left( Q_\theta(s_t, a_t) - \left( r_t + \gamma \mathbb{E}_{s_{t+1} \sim \mathcal{T}} [\overline{V}_{\theta_{\text{targ}}}(s_{t+1})] \right) \right)^2 \right],$$ where $\overline{V}_\chi(s_t) = \mathbb{E}_{a_t \sim \pi} [Q_\chi(s_t, a_t) - \alpha \log \pi(a_t \mid s_t)]$ for a Q-function with parameters $\chi$.

The policy is updated by minimising the expected KL-divergence, $$J_\pi(s_t, \phi) = \mathbb{E}_{s_t \sim D} \left[ \mathbb{E}_{a_t \sim \pi_\phi} [\alpha \log \pi_\phi(a_t \mid s_t) - Q_\theta(s_t, a_t)] \right],$$ where the policy is reparameterised using a neural network transformation, by setting $a_t = f_{\phi}(\epsilon_t; s_t)$ for an input noise vector $\epsilon_t$.

Additionally, the temperature parameter can be automatically adjusted to control the entropy term, $$J(\alpha) = \mathbb{E}_{a_t \sim \pi_t} [-\alpha \log \pi_t(a_t \mid s_t) - \alpha H].$$ By treating the entropy as a constraint, we avoid needing to manually set the often difficult to optimise temperature. The idea here is that the policy should explore regions where the optimal action is uncertain, and become deterministic in regions where the optimal action is clear. $J(\alpha)$ ensures that the average entropy of the policy is constrained, whilst the entropy at different states vary. The temperature eventually decays close to zero during training, making the policy close to fully deterministic.

Full algorithm and gradient details are provided by \citet{haarnoja2019softactorcriticalgorithmsapplications}. We note that SAC employs 2 Q-functions to tackle overestimation bias \citep{fujimoto2018addressing}. The SAC parameter updates explained in the pseudocode below are done similarly for the algorithms that follow this subsection.

\begin{algorithm}[H]
\caption{Soft Actor-Critic}
\begin{algorithmic}[1]
    \STATE Initialise policy parameters $\phi$, Q-function parameters $\theta_1, \theta_2$, and empty replay buffer $D$. Set target parameters $\theta_{\text{targ},1} \leftarrow \theta_1$, $\theta_{\text{targ},2} \leftarrow \theta_2$.
    \FOR{each episode}
        \STATE Take one action $a_t \sim \pi_\phi(\cdot \mid s_t)$. Observe reward $r_t$, new state $s_{t + 1}$.
        \STATE Add data to buffer: $D \leftarrow D \cup \{(s_t, a_t, r_t, s_{t+1})\}$
        \FOR{each gradient step}
            \STATE Sample a mini-batch $B = \{(s_t, a_t, r_t, s_{t+1})\}$ from $D$
            \STATE Compute the Q-target $y$ for the Q-functions by $$y = r_t + \gamma \left(\min_{i = 1, 2} Q_{\theta_{\text{targ},i}}(s_{t + 1}, \tilde{a}_{t + 1}) - \alpha \log \pi_\phi(\tilde{a}_{t + 1} \mid s_{t + 1}) \right), \quad \tilde{a}_{t + 1} \sim \pi_\phi(\cdot \mid s_{t + 1})$$
            \FOR {$i = 1, 2$}
                \STATE \texttt{\# Perform the following update below $\theta_i \gets \theta_i - \lambda_Q \nabla_{\theta_i} \hat{J}_Q(\theta_i)$}
                \STATE Update $\theta_i$ with gradient descent using $$\nabla_{\theta_i} \frac{1}{|B|} \sum_{(s_t, a_t, r_t, s_{t+1}) \in B} \left( Q_{\theta_i}(s_t, a_t) - y \right)^{2}$$
                \STATE Update target networks with $\theta_{\text{targ},i} \leftarrow \tau \theta_{i} + (1 - \tau) \theta_{\text{targ}, i}$
            \ENDFOR
            \STATE \texttt{\# Perform the following update below $\phi \gets \phi - \lambda_\pi \nabla_\phi \hat{J}_\pi(\phi)$}
            \STATE Update policy parameters $\phi$ with gradient ascent using $$\nabla_\phi \frac{1}{|B|} \sum_{s_t \in B} \left( \min_{i=1,2} Q_{\theta_i}(s_t, \tilde{a}_\phi(s_t)) - \alpha \log \pi_\phi(\tilde{a}_\phi(s_t) \mid s_t) \right), \quad \tilde{a}_\phi(s_t) \sim \pi_\phi(\cdot \mid s_t)$$
            \STATE \texttt{\# Perform the following update below $\alpha \gets \alpha - \lambda \nabla_\alpha \hat{J}(\alpha)$}
            \STATE Update temperature parameter $\alpha$ by $$\nabla_\alpha \frac{1}{|B|} \sum_{s_t \in B} \left( - \alpha\log \pi_\phi(\tilde{a}_\phi(s_t) \mid s_t) - \alpha H \right)$$
        \ENDFOR
    \ENDFOR
\end{algorithmic}
\end{algorithm}

\subsection{Randomised Ensembled Double Q-Learning}

REDQ extends SAC through employing various adjustments. This includes a higher UTD ratio (by allowing for $G$ updates as in the algorithm below), the use of an ensemble of Q-functions, and in-target minimisation over a random subset $M$ of the $N$ Q-functions \citep{chenrandomized}.

Here, we can use $M$ Q-functions when computing the Q-target, which makes use of our ensemble of $N$ Q-functions. We could set $M = N$ as in SAC, although here we have the advantage of a large ensemble of Q-functions instead of always having 2 Q-functions. Using a small value of $M$, such as $M = 2$, has been found to work well \citep{chenrandomized}. Following from the SAC algorithm presented above, our Q-target is computed by, $$y = r_t + \gamma \left(\min_{i \in \mathcal{M}} Q_{\theta_{\text{targ},i}}(s_{t + 1}, \tilde{a}_{t + 1}) - \alpha \log \pi_\phi(\tilde{a}_{t + 1} \mid s_{t + 1}) \right), \quad \tilde{a}_{t + 1} \sim \pi_\phi(\cdot \mid s_{t + 1}),$$ where $\mathcal{M}$ is a (random) set of indices of size $M$ for the ensemble of Q-functions. The difference here from SAC is taking $\min_{i \in \mathcal{M}}$ over the Q-target functions instead of $\min_{i = 1, 2}$.

\citet{chenrandomized} find that REDQ maintains stable and near-uniform bias under high UTD ratios due to its ensemble and in-target minimisation components. \citet{blau2022optimizing} likely use REDQ for this reason. Where \citet{chenrandomized} recommend using 10 Q-functions, \citet{blau2022optimizing} instead use 2 Q-functions as in SAC. We assume this choice is due to the computational costs of using a larger ensemble of 10 Q-functions, which we confirm through our empirical tests.

\begin{algorithm}[H]
\caption{Randomised Ensembled Double Q-learning}
\begin{algorithmic}[1]
    \STATE Initialise policy parameters $\phi$, $N$ Q-function parameters $\theta_i$, and empty replay buffer $D$. Set target parameters $\theta_{\text{targ},i} \leftarrow \theta_i$. All for $i = 1, \ldots, N$.
    \FOR{each episode}
        \STATE Take one action $a_t \sim \pi_\phi(\cdot \mid s_t)$. Observe reward $r_t$, new state $s_{t + 1}$.
        \STATE Add data to buffer: $D \leftarrow D \cup \{(s_t, a_t, r_t, s_{t+1})\}$
        \FOR {$G$ updates}
            \STATE Sample a mini-batch $B = \{(s_t, a_t, r_t, s_{t + 1})\}$ from $D$
            \STATE Sample a set $\mathcal{M}$ of $M$ distinct indices from $\{1, 2, \ldots, N\}$
            \STATE Compute the Q-target $y$ (same for all of the $N$ Q-functions) by $$y = r_t + \gamma \left(\min_{i \in \mathcal{M}} Q_{\theta_{\text{targ},i}}(s_{t + 1}, \tilde{a}_{t + 1}) - \alpha \log \pi_\phi(\tilde{a}_{t + 1} \mid s_{t + 1}) \right), \quad \tilde{a}_{t + 1} \sim \pi_\phi(\cdot \mid s_{t + 1})$$
            \FOR {$i = 1, \ldots, N$}
                \STATE Update $\theta_i$ with gradient descent using $$\nabla_{\theta_i} \frac{1}{|B|} \sum_{(s_t,a_t,r_t,s_{t + 1}) \in B} \left( Q_{\theta_i}(s_t, a_t) - y \right)^{2}$$
                \STATE Update target networks with $\theta_{\text{targ},i} \leftarrow \tau \theta_{i} + (1 - \tau) \theta_{\text{targ}, i}$
            \ENDFOR
            \STATE Update policy parameters $\phi$ with gradient ascent using $$\nabla_\phi \frac{1}{|B|} \sum_{s_t \in B} \left( \frac{1}{N} \sum_{i = 1}^{N} Q_{\theta_i}(s_t, \tilde{a}_\phi(s_t)) - \alpha \log \pi_\phi(\tilde{a}_\phi(s_t) \mid s_t) \right), \quad \tilde{a}_\phi(s_t) \sim \pi_\phi(\cdot \mid s_t)$$
            \STATE Update temperature parameter $\alpha$ by $$\nabla_\alpha \frac{1}{|B|} \sum_{s_t \in B} \left( - \alpha\log \pi_\phi(\tilde{a}_\phi(s_t) \mid s_t) - \alpha H \right)$$
        \ENDFOR
    \ENDFOR
\end{algorithmic}
\end{algorithm}

\subsection{Dropout Q-Functions for Doubly Efficient Reinforcement Learning}

DroQ is an algorithm that seeks to improve on the computational costs of REDQ by making use of both dropout regularisation \citep{hinton2012improving} and layer normalisation \citep{ba2016layer} in the deep Q-functions. A much smaller ensemble of Q-functions can be used due to the inclusion of dropout, such as $N = 2$. The main difference here is the use of dropout Q-functions, which we denote by $Q_{\text{Dr}}$ in the algorithm below. There is no in-target minimisation as in REDQ, so we compute the Q-targets using every Q-function in the ensemble.

\begin{algorithm}[H]
\caption{Dropout Q-Functions for Doubly Efficient Reinforcement Learning}
\begin{algorithmic}[1]
    \STATE Initialise policy parameters $\phi$, $N$ Q-function parameters $\theta_i$ and empty replay buffer $D$. Set target parameters $\theta_{\text{targ},i} \leftarrow \theta_i$. All for $i = 1, \ldots, N$.
    \FOR{each episode}
        \STATE Take action $a_t \sim \pi_\phi(\cdot \mid s_t)$. Observe reward $r_t$, next state $s_{t+1}$.
        \STATE $D \leftarrow D \cup \{(s_t, a_t, r_t, s_{t+1})\}$.
        \FOR{$G$ updates}
            \STATE Sample a mini-batch $B = \{(s_t, a_t, r_t, s_{t + 1})\}$ from $D$.
            \STATE Compute the Q-target $y$ for the dropout Q-functions by $$y = r_t + \gamma \left( \min_{i=1,\ldots,N} Q_{\text{Dr},\theta_{\text{targ},i}}(s_{t + 1}, \tilde{a}_{t + 1}) - \alpha \log \pi_\phi(\tilde{a}_{t + 1} \mid s_{t + 1}) \right), \quad \tilde{a}_{t + 1} \sim \pi_\phi(\cdot \mid s_{t + 1})$$
            \FOR{$i = 1, \ldots, N$}
                \STATE Update $\theta_i$ with gradient descent using $$\nabla_{\theta_i} \frac{1}{|B|} \sum_{(s_t,a_t,r_t,s_{t + 1}) \in B} \left( Q_{\text{Dr},\theta_i}(s_t, a_t) - y \right)^{2}$$
                \STATE Update target networks with $\theta_{\text{targ},i} \leftarrow \tau \theta_{i} + (1 - \tau) \theta_{\text{targ}, i}$
            \ENDFOR
            \STATE Update policy parameters $\phi$ with gradient ascent using $$\nabla_\phi \frac{1}{|B|} \sum_{s_t \in B} \left( \frac{1}{N} \sum_{i = 1}^{N} Q_{\text{Dr},\theta_i}(s_t, \tilde{a}_\phi(s_t)) - \alpha \log \pi_\phi(\tilde{a}_\phi(s_t) \mid s_t) \right), \quad \tilde{a}_\phi(s_t) \sim \pi_\phi(\cdot \mid s_t)$$
            \STATE Update temperature parameter $\alpha$ by $$\nabla_\alpha \frac{1}{|B|} \sum_{s_t \in B} \left( - \alpha\log \pi_\phi(\tilde{a}_\phi(s_t) \mid s_t) - \alpha H \right)$$
        \ENDFOR
    \ENDFOR
\end{algorithmic}
\end{algorithm}

\subsection{Scaled-by-Resetting}

SBR \citep{doro2022sampleefficient} explores how increasing the replay ratio, the number of times an agent's parameters are updated per environment interaction, can drastically improve sample efficiency. Essentially, the parameters of the neural networks for the agent are periodically reset, either partially or fully, during training. It is fully reset for SAC \citep{doro2022sampleefficient}, and so we employ the same in our work.

In more detail, the Q-functions and policy networks are reset, meaning that the neural network parameters are back to their untrained state. We only retain the replay buffer, and after a reset, the agent is again trained as normal, this time updating its parameters using the previously gathered experiences stored in the replay buffer. This is close to the offline RL setting, where a dataset of experiences is already made available, though here we are also collecting experiences online at the same time.

Our implementation of SBR is built over REDQ, as seen in the pseudocode below.

\begin{algorithm}[H]
\caption{Scaled-by-Resetting}
\begin{algorithmic}[1]
    \STATE Initialise policy parameters $\phi$, $N$ Q-function parameters $\theta_i$, and empty replay buffer $D$. Set target parameters $\theta_{\text{targ},i} \leftarrow \theta_i$. All for $i = 1, \ldots, N$. Set fixed reset interval $\psi$.
    \FOR{each episode}
        \STATE Take one action $a_t \sim \pi_\phi(\cdot \mid s_t)$. Observe reward $r_t$, new state $s_{t + 1}$.
        \STATE Add data to buffer: $D \leftarrow D \cup \{(s_t, a_t, r_t, s_{t+1})\}$
        \FOR {$G$ updates}
            \STATE Sample a mini-batch $B = \{(s_t, a_t, r_t, s_{t+1})\}$ from $D$
            \STATE Sample a set $\mathcal{M}$ of $M$ distinct indices from $\{1, 2, \ldots, N\}$
            \STATE Compute the Q-target $y$ (same for all of the $N$ Q-functions) by $$y = r_t + \gamma \left(\min_{i \in \mathcal{M}} Q_{\theta_{\text{targ},i}}(s_{t + 1}, \tilde{a}_{t + 1}) - \alpha \log \pi_\phi(\tilde{a}_{t + 1} \mid s_{t + 1}) \right), \quad \tilde{a}_{t + 1} \sim \pi_\phi(\cdot \mid s_{t + 1})$$
            \FOR {$i = 1, \ldots, N$}
                \STATE Update $\theta_i$ with gradient descent using $$\nabla_{\theta_i} \frac{1}{|B|} \sum_{(s_t,a_t,r_t,s_{t + 1}) \in B} \left( Q_{\theta_i}(s_t, a_t) - y \right)^{2}$$
                \STATE Update target networks with $\theta_{\text{targ},i} \leftarrow \tau \theta_{i} + (1 - \tau) \theta_{\text{targ}, i}$
            \ENDFOR
            \STATE Update policy parameters $\phi$ with gradient ascent using $$\nabla_\phi \frac{1}{|B|} \sum_{s_t \in B} \left( \frac{1}{N} \sum_{i = 1}^{N} Q_{\theta_i}(s_t, \tilde{a}_\phi(s_t)) - \alpha \log \pi_\phi(\tilde{a}_\phi(s_t) \mid s_t) \right), \quad \tilde{a}_\phi(s_t) \sim \pi_\phi(\cdot \mid s_t)$$
            \STATE Update temperature parameter $\alpha$ by $$\nabla_\alpha \frac{1}{|B|} \sum_{s_t \in B} \left( - \alpha\log \pi_\phi(\tilde{a}_\phi(s_t) \mid s_t) - \alpha H \right)$$
            \IF{number of gradient steps exceeds $\psi$}
                \STATE Reset $\phi$, $\theta_i, \ldots \theta_N$, $\theta_{\text{targ},1}, \ldots, \theta_{\text{targ},N}$ as done initially, keeping the replay buffer
                \STATE $\psi \leftarrow \psi + \psi$
            \ENDIF
        \ENDFOR
    \ENDFOR
\end{algorithmic}
\end{algorithm}

\subsection{Simple Unified Framework for Reinforcement Learning Using Ensembles}

SUNRISE is a method designed to enhance off-policy RL algorithms by addressing common challenges like instability in Q-learning \citep{lee2021sunrise}. It achieves this by integrating ensemble-based weighted Bellman backups, which re-weight target Q-values based on uncertainty estimates from a Q-ensemble, and a UCB exploration strategy that selects actions with the highest UCB to encourage efficient exploration.

We explain the UCB exploration strategy \citep{chen2017ucbexplorationqensembles, lee2021sunrise} in full, which is an important component of our policy ensemble-based algorithm. The UCB exploration formula used in the SUNRISE framework \citep{lee2021sunrise} is designed to balance exploration and exploitation through using the $N$ candidate actions generated by our $N$ policies. The UCB formula is as follows, $$a_{t} = \argmax\limits_{a} \{Q_{\text{mean}}(s_{t}, a) + \lambda Q_{\text{std}}(s_{t}, a)\}.$$

In this formula, $a_{t}$ is the action selected at timestep $t$, which maximises the sum of the mean and a scaled standard deviation of the Q-values. \(Q_{\text{mean}}(s_{t}, a)\) represents the average estimated value of taking action \(a\) in state $s_{t}$, derived from an ensemble of Q-functions. $Q_{\text{std}}(s_{t}, a)$ denotes the uncertainty or variance in these Q-value estimates. $\lambda$ controls the weight of the uncertainty term, thus modulating the level of exploration.

Consider an ensemble of $N$ SAC agents, $\{Q_{\theta_{i}}, \pi_{\phi_{i}}\}_{i = 1}^{N}$, where $\theta_{i}$ and $\phi_{i}$ denote the parameters of the $i$-th soft Q-function and policy, respectively. Conventional Q-learning is based on the Bellman backup, and so it can be affected by error propagation. This essentially means that errors in the previous Q-function induce ``noise" to the true Q-value of the current Q-function. Therefore, for each agent $i$, a weighted Bellman backup \citep{lee2021sunrise} is used instead of $J_{Q}$ from SAC \citep{haarnoja2019softactorcriticalgorithmsapplications}, given by, $$J_{WQ}(\tau_t, \theta_i) = w (s_{t + 1}, a_{t + 1}) \left( Q_{\theta_i}(s_t, a_t) - r_t - \gamma \overline{V}_{\theta_{\text{targ},i}}(s_{t + 1}) \right)^{2},$$
where $\tau_t = (s_t, a_t, r_t, s_{t + 1})$ is a transition, $a_{t + 1} \sim \pi_{\phi_i}(a_t \mid s_t)$, $\overline{V}$ is the value function, defined as in SAC, and $w(s_t, a_t)$ is a confidence weight based on an ensemble of target Q-functions. 

The confidence weight $w(s_t, a_t)$ \citep{lee2021sunrise} is defined as, $$w(s_t, a_t) = \sigma \left( -\overline{Q}_{\text{std}}(s_t, a_t) \cdot \delta \right) + 0.5,$$ where $\delta > 0$ is a temperature, $\sigma$ is the sigmoid function, and $\overline{Q}_{\text{std}}(s_t, a_t)$ is the empirical standard deviation of all target Q-functions $\{ Q_{\theta_{\text{targ},i}} \}_{i = 1}^{N}$.

\begin{algorithm}[H]
\caption{Simple Unified Framework for Reinforcement Learning Using Ensembles}
\begin{algorithmic}[1]
    \STATE Initialise $N$ policy parameters $\phi_i$, $N$ Q-function parameters $\theta_i$, and empty replay buffer $D$. Set target parameters $\theta_{\text{targ},i} \leftarrow \theta_i$ and initial temperature parameters $\alpha_i$. All for $i = 1, \ldots, N$.
    \FOR{each episode}
        \FOR{each timestep $t$}
            \STATE Collect $N$ action samples: $A_t = \{a_{t, i} \sim \pi_{\phi_i}(a \mid s_t) \mid i \in \{1, \ldots, N\}\}$
            \STATE Choose the action that maximises UCB by $$a_t = \argmax\limits_{a_{t, i} \in A_t} \left( Q_{\text{mean}}(s_t, a_{t, i}) + \lambda Q_{\text{std}}(s_t, a_{t,i}) \right)$$
            \STATE Collect state $s_{t + 1}$ and reward $r_t$ from the environment by taking action $a_t$
            \STATE Sample bootstrap masks $M_t = \{m_{t, i} \sim \text{Bernoulli}(\beta) \mid i \in \{1, \ldots, N\}\}$
            \STATE Store transitions $\tau_t = (s_t, a_t, s_{t + 1}, r_t)$ and masks in replay buffer $D \leftarrow D \cup \{(\tau_t, M_t)\}$
        \ENDFOR
        \FOR{$G$ updates}
            \STATE Sample random minibatch $\{(\tau_j, M_j)\}_{j=1}^B \sim D$
            \FOR{each agent $i$}
                \STATE \texttt{\# $J_{WQ}$ as found above}
                \STATE Update the Q-function by minimising $$\nabla_{\theta_i}\frac{1}{|B|} \sum_{j=1}^B m_{j,i} \cdot J_{WQ}(\tau_j, \theta_i)$$
                \STATE Update target networks with $\theta_{\text{targ},i} \leftarrow \tau \theta_{i} + (1 - \tau) \theta_{\text{targ}, i}$
                \STATE \texttt{\# $J_\pi$ as found in SAC}
                \STATE Update the policy by minimising $$\nabla_{\phi_i}\frac{1}{|B|} \sum_{j=1}^B m_{j,i} \cdot J_\pi(s_j, \phi_i)$$
                \STATE \texttt{\# $J(\alpha)$ as found in SAC}
                \STATE Update temperature parameter by minimising $$\nabla_{\alpha_i}\frac{1}{|B|} \sum_{j=1}^B m_{j,i} \cdot J(\alpha_i)$$
            \ENDFOR
        \ENDFOR
    \ENDFOR
\end{algorithmic}
\end{algorithm}

Our implementation of SUNRISE slightly differs from the default algorithm above, where the only difference is that we do not sample any bootstrap masks, which is simply the equivalent of $\beta = 1$ (where the binary masks would all be equal to 1). \citet{lee2021sunrise} find that $\beta = 1$ performs the best, making this a sensible adjustment to the algorithm that can also reduce the overall agent training time. See Appendix F for how we decided to deploy our SUNRISE agents at deployment time.

\section{\raggedright{Training Regimes}}

In this section, we explain our training regimes used to obtain our results in more detail. This includes our exact neural network architectures and any hyperparameter optimisation.


\subsection{Neural Network, Training, Environment, and Evaluation Details}

\subsubsection{Neural Networks}

In all of our experiments, the (trained) policies are neural networks that consist of an encoder network and an emitter network, as we explain in Appendix A.

Our encoder network, which takes the observation space dimensions as its input, consists of two hidden layers, each with 128 nodes, and the rectified linear unit (ReLU) activation function is used to introduce nonlinearity between the layers. The final pooled representation from the encoder consists of $\frac{128}{2} = 64$ nodes, without any activation function. Our emitter network, which takes as input the encoded representation from the encoder, consists of two hidden layers, each with 128 nodes, and the rectified linear unit (ReLU) activation function is used to introduce nonlinearity between the layers. This is evaluated as usual as a TanhNormal policy in SAC \citep{haarnoja2019softactorcriticalgorithmsapplications}, and we obtain a (stochastic) distribution over the design space as output.

The Q-functions follow the same structure as the policies, and depending on whether dropout and layer normalisation are used, they have the same architecture in every experiment. If dropout and layer normalisation are required, they are deployed before the activation function, starting with dropout. The Q-functions each output a Q-value for all actions.

\subsubsection{Training}
We use the Adam \citep{kingma2017adammethodstochasticoptimization} optimiser for the policy, the Q-functions, and for controlling $\alpha$ (the exploration parameter) during training. All agents are trained using 20000 iterations. We train with $L = 1e5$ contrastive samples for every agent on the sPCE objective. We sample 4096 transitions from the replay buffer for a single optimisation step (mini-batch size). 10 agents are trained under unique random seeds for each algorithm, which we use for our evaluation later.

\subsubsection{Environment}

In RL, it is possible to normalise/scale the actions, observations, and rewards in the environment (our experimental design problem). Since we want the exact sPCE, we do not perform reward scaling. Actions are scaled to be in range [-1, 1], and observations are scaled to be in range [0, 1].

\subsubsection{Evaluation}

RL is known to not be very robust to random seeds, so measuring the average performance across agents trained with unique random seeds is a sensible approach to statistically evaluating our algorithms (both during training and at deployment time) \citep{colas2022hitchhikersguidestatisticalcomparisons}. At deployment/evaluation time, we take the average performance of each agent across 2000 rollouts, meaning 2000 different choices of $\boldsymbol{\theta}$ in our Bayesian experimental design problem. Bear in mind that this is for a single agent that has been trained with a particular random seed, and so we do the same for all 10 agents trained with unique random seeds, and take the average performance of this. Since we have 10 agents, our final average performance for each metric would be based on $2000 \cdot 10 = 20000$ different choices of $\boldsymbol{\theta}$ in our experimental design problem. This is a reasonably large number of scenarios to consider, much larger than that by \citet{blau2022optimizing}, making our results robust.

\subsection{REDQ Hyperparameters}

\citet{blau2022optimizing} perform some hyperparameter optimisation on the target update rate $\tau$, policy learning rate, critic (or Q-function) learning rate, replay buffer size, and discount factor $\gamma$. Their best set of hyperparameters are listed in the table below as `REDQ'. This is the same set of hyperparameters we used for the results explained in the main body of this paper, for a fair comparison to \citet{blau2022optimizing}.

We additionally test on our own accord a discount factor of 0.99 (REDQ-Disc-0.99), ensemble of 10 Q-functions as done by \citet{chenrandomized} (REDQ-Ens-10), and a target update rate of 0.01 (REDQ-Tau-0.01).

\begin{table}[H]
    \centering
    \begin{tabular}{ccccc}
    \hline
    Parameter & REDQ & REDQ-Disc-0.99 & REDQ-Ens-10 & REDQ-Tau-0.01 \\
    \hline
    $N$ & 2 & 2 & 10 & 2 \\
    $M$ & 2 & 2 & 2 & 2 \\
    $\gamma$ & 0.9 & 0.99 & 0.9 & 0.9 \\
    $\tau$ & 0.001 & 0.001 & 0.001 & 0.01 \\
    UTD Ratio $G$ & 64 & 64 & 64 & 64 \\
    Policy Learning Rate & 0.0001 & 0.0001 & 0.0001 & 0.0001 \\
    Critic Learning Rate & 0.0003 & 0.0003 & 0.0003 & 0.0003 \\
    Buffer Size & 10000000 & 10000000 & 10000000 & 10000000 \\
    \end{tabular}
    \caption{REDQ Hyperparameter Combinations}
\end{table}

The sPCE and sNMC results at deployment time from using these hyperparameters are explained in the tables below:

\begin{table}[H]
    \centering
    \begin{tabular}{cccccccc}
         & \multicolumn{5}{c}{sPCE} & Time \\
        \hline
        & $K = 1$ & $K = 2$ & $K = 3$ & $K = 4$ & $K = 5$ & $K = 2$ \\
        \hline
        REDQ & $6.279 \pm 0.013$ & $11.689 \pm 0.012$ & $11.881 \pm 0.013$ & $11.507 \pm 0.015$ & $11.095 \pm 0.017$ & \textbf{13.23h} \\
        REDQ-Disc-0.99 & $6.328 \pm 0.012$ & $11.536 \pm 0.013$ & $11.685 \pm 0.014$ & $11.250 \pm 0.015$ & $10.825 \pm 0.017$ & 13.41h \\
        REDQ-Ens-10 & \textbf{6.423 $\pm$ 0.013} & \textbf{11.744 $\pm$ 0.013} & $11.884 \pm 0.013$ & $11.419 \pm 0.015$ & $10.989 \pm 0.017$ & 38.91h \\
        REDQ-Tau-0.01 & $6.286 \pm 0.013$ & $11.651 \pm 0.013$ & \textbf{11.919 $\pm$ 0.013} & \textbf{11.516 $\pm$ 0.015} & \textbf{11.177 $\pm$ 0.016}& 13.38h
    \end{tabular}
    \caption{sPCE at $T = 30$ computed using $L = 1e6$, and average agent training time for the location finding experiment. Means and standard errors are from 20000 rollouts, spread evenly across 10 agents trained under unique random seeds. The average training time is across these 10 agents.}
    \label{spce1}
\end{table}

\begin{table}[H]
    \centering
    \begin{tabular}{cccccccc}
         & \multicolumn{5}{c}{sNMC} \\
        \hline
        & $K = 1$ & $K = 2$ & $K = 3$ & $K = 4$ & $K = 5$\\
        \hline
        REDQ & $6.282 \pm 0.013$ & $12.323 \pm 0.019$ & $13.215 \pm 0.029$ & $12.726 \pm 0.030$ & $12.190 \pm 0.030$ \\
        REDQ-Disc-0.99 & $6.331 \pm 0.012$ & $12.071 \pm 0.018$ & $12.704 \pm 0.025$ & $12.170 \pm 0.027$ & $11.653 \pm 0.028$ \\
        REDQ-Ens-10 & \textbf{6.425 $\pm$ 0.013} & \textbf{12.427 $\pm$ 0.019} & $13.140 \pm 0.028$ & $12.513 \pm 0.029$ & $11.943 \pm 0.029$ \\
        REDQ-Tau-0.01 & $6.289 \pm 0.013$ & $12.270 \pm 0.019$ & \textbf{13.261 $\pm$ 0.028} & \textbf{12.755 $\pm$ 0.030} & \textbf{12.332 $\pm$ 0.031}
    \end{tabular}
    \label{snmc1}
    \caption{sNMC at $T = 30$ computed using $L = 1e6$ for the location finding experiment. Means and standard errors are from 20000 rollouts, spread evenly across 10 agents trained under unique random seeds.}
\end{table}

For the location finding experiment, REDQ-Ens-10 is the most expensive to train, and only provides the best results for $K \in \{1, 2\}$. REDQ-Tau-0.01 provides the best results for $K \geq 3$, in a much cheaper amount of time.

\begin{table}[H]
    \centering
    \begin{tabular}{cccccccc}
         & \multicolumn{2}{c}{sPCE} & \multicolumn{2}{c}{sNMC} & Time \\
        \hline
        & $\nu = 0.005$ & $\nu = 0.01$ & $\nu = 0.005$ & $\nu = 0.01$ & $\nu = 0.005$ \\
        \hline
        REDQ-Blau & \textbf{13.782 $\pm$ 0.021} & $12.343 \pm 0.022$ & $19.911 \pm 0.155$ & $13.433 \pm 0.046$ & \textbf{8.95h} \\
        REDQ-Disc-0.99 & $13.723 \pm 0.021$ & $12.317 \pm 0.023$ & $20.240 \pm 0.164$ & $13.494 \pm 0.049$ & 10.69h \\
        REDQ-Ens-10 & $13.781 \pm 0.021$ & $12.304 \pm 0.021$ & $18.989 \pm 0.138$ & $13.192 \pm 0.042$ & 24.75h \\
        REDQ-Tau-0.01 & $13.763 \pm 0.022$ & \textbf{12.413 $\pm$ 0.023} & \textbf{22.116 $\pm$ 0.216} & \textbf{14.034 $\pm$ 0.061} & 10.67h
    \end{tabular}
    \caption{sPCE and sNMC at $T = 10$ computed using $L = 1e7$, and average agent training time for the CES experiment. Means and standard errors are from 20000 rollouts, spread evenly across 10 agents trained under unique random seeds. The average training time is across these 10 agents.}
\end{table}

For the CES experiment, REDQ-Blau is best in terms of sPCE when evaluated on the same statistical model used during training. It falls behind REDQ-Tau-0.01 when the model is changed. REDQ-Tau-0.01 provides the greatest sNMC values, suggesting that $\tau = 0.01$ may be sensible to train agents with.


\subsection{DroQ Hyperparameters}

\citet{hiraoka2022dropout} find empirically that smaller dropout rates provide better performance over larger ones. We compare the performance between dropout rates of 0.01 and 0.1. There are no resets for DroQ and so these are left blank in Table 7 below:

\begin{table}[H]
    \centering
    \begin{tabular}{ccccccc}
    \hline
    Parameter & DroQ-0.01 & DroQ-0.1 & SBR-300000 & SBR-430000 \\
    \hline
    $N$ & 2 & 2 & 2 & 2 \\
    $M$ & 2 & 2 & 2 & 2 \\
    $\gamma$ & 0.9 & 0.9 & 0.9 & 0.9 \\
    $\tau$ & 0.001 & 0.001 & 0.001 & 0.001 \\
    UTD Ratio $G$ & 64 & 64 & 64 & 64 \\
    Dropout Probability & 0.01 & 0.1 & - & - \\
    Reset Interval & - & - & 300000 & 430000 \\
    Policy Learning Rate & 0.0001 & 0.0001 & 0.0001 & 0.0001 \\
    Critic Learning Rate & 0.0003 & 0.0003 & 0.0003 & 0.0003 \\
    Buffer Size & 10000000 & 10000000 & 10000000 & 10000000 \\
    \end{tabular}
    \caption{DroQ and SBR Hyperparameter Combinations}
\end{table}

For the location finding finding experiment we use 0.01, and for the CES experiment we use 0.1. These are the best hyperparameter combinations as explained in the tables below:

\begin{table}[H]
    \centering
    \begin{tabular}{cccccccc}
         & \multicolumn{5}{c}{sPCE} & Time \\
        \hline
        & $K = 1$ & $K = 2$ & $K = 3$ & $K = 4$ & $K = 5$ & $K = 2$ \\
        \hline
        DroQ-0.01 & \textbf{6.368 $\pm$ 0.013} & \textbf{11.680 $\pm$ 0.013} & \textbf{11.902 $\pm$ 0.013} & \textbf{11.480 $\pm$ 0.015} & \textbf{11.090 $\pm$ 0.017} & \textbf{19.05h} \\
        DroQ-0.1 & $6.333 \pm 0.013$ & $11.176 \pm 0.013$ & $11.708 \pm 0.014$ & $11.384 \pm 0.016$ & $11.090 \pm 0.017$ & 19.63h
    \end{tabular}
    \label{spce2}
    \caption{sPCE at $T = 30$ computed using $L = 1e6$, and average agent training time for the location finding experiment. Means and standard errors are from 20000 rollouts, spread evenly across 10 agents trained under unique random seeds. The average training time is across these 10 agents.}
\end{table}

\begin{table}[H]
    \centering
    \begin{tabular}{cccccccc}
         & \multicolumn{5}{c}{sNMC} \\
        \hline
        & $K = 1$ & $K = 2$ & $K = 3$ & $K = 4$ & $K = 5$ \\
        \hline
        DroQ-0.01 & \textbf{6.371 $\pm$ 0.013} & \textbf{12.349 $\pm$ 0.020} & \textbf{13.231 $\pm$ 0.028} & \textbf{12.669 $\pm$ 0.029} & $12.147 \pm 0.030$ \\
        DroQ-0.1 & $6.336 \pm 0.013$ & $11.626 \pm 0.019$ & $12.980 \pm 0.029$ & $12.594 \pm 0.030$ & \textbf{12.232 $\pm$ 0.031}
    \end{tabular}
    \label{snmc2}
    \caption{sNMC at $T = 30$ computed using $L = 1e6$ for the location finding experiment. Means and standard errors are from 20000 rollouts, spread evenly across 10 agents trained under unique random seeds.}
\end{table}

\begin{table}[H]
    \centering
    \begin{tabular}{cccccccc}
         & \multicolumn{2}{c}{sPCE} & \multicolumn{2}{c}{sNMC} & Time \\
        \hline
        & $\nu = 0.005$ & $\nu = 0.01$ & $\nu = 0.005$ & $\nu = 0.01$ & $\nu = 0.005$ \\
        \hline
        DroQ-0.01 & $13.902 \pm 0.020$ & $12.427 \pm 0.021$ & $19.904 \pm 0.140$ & $13.479 \pm 0.043$ & \textbf{12.65h} \\
        DroQ-0.1 & \textbf{14.090 $\pm$ 0.020} & \textbf{12.683 $\pm$ 0.022} & \textbf{21.764 $\pm$ 0.201} & \textbf{14.108 $\pm$ 0.058} & 13.18h
    \end{tabular}
    \caption{sPCE and sNMC at $T = 10$ computed using $L = 1e7$, and average agent training time for the CES experiment. Means and standard errors are from 20000 rollouts, spread evenly across 10 agents trained under unique random seeds. The average training time is across these 10 agents.}
\end{table}


Using a dropout rate of 0.01 is clearly more competitive than using a dropout rate of 0.1 for the location finding experiment. This is reversed for the CES experiment.

\subsection{SBR Hyperparameters}

One hyperparameter set looks at 2 resets of the neural network parameters during training, and the other looks at 4 resets during training. These are represented as reset intervals $x$ in the algorithm, where we perform a reset every $x$ amount of gradient steps. Table 7 displays the SBR hyperparameter combinations. There is no dropout involved, and so these are left blank (or equivalently set to 0).


We use SBR-430000 for both experimental design problems, due to the explained performance provided in the tables below:

\begin{table}[H]
    \centering
    \begin{tabular}{cccccccc}
         & \multicolumn{5}{c}{sPCE} & Time \\
        \hline
        & $K = 1$ & $K = 2$ & $K = 3$ & $K = 4$ & $K = 5$ & $K = 2$ \\
        \hline
        SBR-300000 & $6.020 \pm 0.012$ & $11.178 \pm 0.013$ & $11.773 \pm 0.014$ & \textbf{11.608 $\pm$ 0.015} & \textbf{11.436 $\pm$ 0.016} & \textbf{13.30h} \\
        SBR-430000 & \textbf{6.165 $\pm$ 0.012} & \textbf{11.362 $\pm$ 0.013} & \textbf{11.788 $\pm$ 0.014} & $11.518 \pm 0.016$ & $11.257 \pm 0.017$ & 13.52h
    \end{tabular}
    \label{spce3}
    \caption{sPCE at $T = 30$ computed using $L = 1e6$, and average agent training time for the location finding experiment. Means and standard errors are from 20000 rollouts, spread evenly across 10 agents trained under unique random seeds. The average training time is across these 10 agents.}
\end{table}

\begin{table}[H]
    \centering
    \begin{tabular}{cccccccc}
         & \multicolumn{5}{c}{sNMC} \\
        \hline
        & $K = 1$ & $K = 2$ & $K = 3$ & $K = 4$ & $K = 5$ \\
        \hline
        SBR-300000 & $6.021 \pm 0.012$ & $11.515 \pm 0.016$ & $12.960 \pm 0.027$ & \textbf{12.991 $\pm$ 0.031} & \textbf{12.826 $\pm$ 0.032} \\
        SBR-430000 & \textbf{6.168 $\pm$ 0.013} & \textbf{11.793 $\pm$ 0.017} & \textbf{13.043 $\pm$ 0.028} & $12.886 \pm 0.032$ & $12.571 \pm 0.032$
    \end{tabular}
    \label{snmc3}
    \caption{sNMC at $T = 30$ computed using $L = 1e6$ for the location finding experiment. Means and standard errors are from 20000 rollouts, spread evenly across 10 agents trained under unique random seeds.}
\end{table}

For the location finding experiment, we do note that the results for $K \geq 4$ suggest that SBR-300000 is better for this setting. SBR-430000 is otherwise the strongest performer for $K < 4$, which is the reason why we choose this as our main hyperparameter combination.

\begin{table}[H]
    \centering
    \begin{tabular}{cccccccc}
         & \multicolumn{2}{c}{sPCE} & \multicolumn{2}{c}{sNMC} & Time \\
        \hline
        & $\nu = 0.005$ & $\nu = 0.01$ & $\nu = 0.005$ & $\nu = 0.01$ & $\nu = 0.005$ \\
        \hline
        SBR-300000 & $13.047 \pm 0.026$ & $11.688 \pm 0.026$ & $18.248 \pm 0.134$ & $12.586 \pm 0.044$ & \textbf{9.02h} \\
        SBR-430000 & \textbf{13.640 $\pm$ 0.022} & \textbf{12.241 $\pm$ 0.023} & \textbf{20.567 $\pm$ 0.179} & \textbf{13.519 $\pm$ 0.052} & 9.03h
    \end{tabular}
    \caption{sPCE and sNMC at $T = 10$ computed using $L = 1e7$, and average agent training time for the CES experiment. Means and standard errors are from 20000 rollouts, spread evenly across 10 agents trained under unique random seeds. The average training time is across these 10 agents.}
\end{table}


SBR-430000 performs best for both sPCE and sNMC on the two statistical models considered.

\subsection{SUNRISE Hyperparameters}

We perform hyperparameter optimisation on the Bellman temperature $\delta$, rather than on the UCB weight $\lambda$. To avoid heavy computation, we stick to $N = 2$ agents, making the algorithm comparable to the others explored. We explore $\delta = 10$ (SUNRISE-10) and $\delta = 20$ (SUNRISE-20). There is no dropout involved, and so these are left blank (or equivalently set to 0).

\begin{table}[H]
    \centering
    \begin{tabular}{ccccc}
    \hline
    Parameter & SUNRISE-10 & SUNRISE-20 & Ours-Location & Ours-CES \\
    \hline
    $N$ & 2 & 2 & 2 & 2 \\
    $\gamma$ & 0.9 & 0.9 & 0.9 & 0.9 \\
    $\tau$ & 0.001 & 0.001 & 0.01 & 0.01 \\
    UTD Ratio $G$ & 64 & 64 & 64 & 64 \\
    UCB $\lambda$ & 1 & 1 & 1 & 1 \\
    Dropout Probability & - & - & 0.01 & 0.1 \\
    Bellman Temperature $\delta$ & 10 & 20 & 20 & 10 \\
    Policy Learning Rate & 0.0001 & 0.0001 & 0.0001 & 0.0001 \\
    Critic Learning Rate & 0.0003 & 0.0003 & 0.0003 & 0.0003 \\
    Buffer Size & 10000000 & 10000000 & 10000000 & 10000000 \\
    \end{tabular}
    \caption{SUNRISE and SUNRISE-DroQ Hyperparameter Combinations}
\end{table}

We use SUNRISE-20 for the location finding experiment and SUNRISE-10 for the CES experiment, due to the results explained in the tables below:

\begin{table}[H]
    \centering
    \begin{tabular}{cccccccc}
         & \multicolumn{5}{c}{sPCE} & Time \\
        \hline
        & $K = 1$ & $K = 2$ & $K = 3$ & $K = 4$ & $K = 5$ & $K = 2$ \\
        \hline
        SUNRISE-10 & \textbf{6.355 $\pm$ 0.012} & $11.782 \pm 0.012$ & $12.094 \pm 0.013$ & $11.837 \pm 0.014$ & $11.407 \pm 0.016$ & \textbf{21.17h} \\
        SUNRISE-20 & $6.340 \pm 0.013$ & \textbf{11.837 $\pm$ 0.012} & \textbf{12.133 $\pm$ 0.013} & \textbf{11.846 $\pm$ 0.014} & \textbf{11.445 $\pm$ 0.016} & 21.44h \\
    \end{tabular}
    \label{spce4}
    \caption{sPCE at $T = 30$ computed using $L = 1e6$, and average agent training time for the location finding experiment. Means and standard errors are from 20000 rollouts, spread evenly across 10 agents trained under unique random seeds. The average training time is across these 10 agents.}
\end{table}

\begin{table}[H]
    \centering
    \begin{tabular}{cccccccc}
         & \multicolumn{5}{c}{sNMC} \\
        \hline
        & $K = 1$ & $K = 2$ & $K = 3$ & $K = 4$ & $K = 5$ \\
        \hline
        SUNRISE-10 & \textbf{6.358 $\pm$ 0.012} & $12.477 \pm 0.019$ & \textbf{13.662 $\pm$ 0.030} & \textbf{13.397 $\pm$ 0.032} & $12.765 \pm 0.032$ \\
        SUNRISE-20 & $6.342 \pm 0.013$ & \textbf{12.555 $\pm$ 0.019} & $13.656 \pm 0.029$ & $13.345 \pm 0.030$ & \textbf{12.823 $\pm$ 0.032} \\
    \end{tabular}
    \label{snmc4}
    \caption{sNMC at $T = 30$ computed using $L = 1e6$ for the location finding experiment. Means and standard errors are from 20000 rollouts, spread evenly across 10 agents trained under unique random seeds.}
\end{table}

SUNRISE-20 performs best for all $K$, bar $K = 1$ on the location finding experiment in terms of sPCE. We note that the sNMC values are higher for SUNRISE-10 under certain values of $K$. Since sPCE is our main metric of interest, ensuring that the true EIG is at least equal to or higher than a certain value, we choose to stick with SUNRISE-20.

\begin{table}[H]
    \centering
    \begin{tabular}{cccccccc}
         & \multicolumn{2}{c}{sPCE} & \multicolumn{2}{c}{sNMC} & Time \\
        \hline
        & $\nu = 0.005$ & $\nu = 0.01$ & $\nu = 0.005$ & $\nu = 0.01$ & $\nu = 0.005$ \\
        \hline
        SUNRISE-10 & \textbf{13.725 $\pm$ 0.022} & \textbf{12.381 $\pm$ 0.023} & \textbf{21.429 $\pm$ 0.199} & \textbf{13.812 $\pm$ 0.058} & \textbf{15.11h} \\
        SUNRISE-20 & $13.633 \pm 0.022$ & $12.261 \pm 0.023$ & $19.815 \pm 0.130$ & $13.344 \pm 0.042$ & 15.20h \\
    \end{tabular}
    \caption{sPCE and sNMC at $T = 10$ computed using $L = 1e7$, and average agent training time for the CES experiment. Means and standard errors are from 20000 rollouts, spread evenly across 10 agents trained under unique random seeds. The average training time is across these 10 agents.}
\end{table}


For the CES experiment, SUNRISE-10 stands out as the best performer across all evaluated statistical models and metrics.

\subsection{SUNRISE with Dropout Q-Functions Hyperparameters}

There is no hyperparameter optimisation involved for SUNRISE-DroQ. We use our own intuition for selecting the hyperparameters to train with based on the results from the previous algorithms. Our final chosen hyperparameters are presented in Table 14, where Ours-Location is the set used for the location finding experiment, and Ours-CES is the set used for the CES experiment.


The following tables are for Ours-Location and Ours-CES, respectively (note that the deployment sPCE and sNMC results below are for the same trained agents, so the training time displayed is equal):

\begin{table}[H]
    \centering
    \begin{tabular}{cccccccc}
         & \multicolumn{5}{c}{} & Time \\
        \hline
        & $K = 1$ & $K = 2$ & $K = 3$ & $K = 4$ & $K = 5$ & $K = 2$ \\
        \hline
        Ours-sPCE & $6.433 \pm 0.012$ & $11.770 \pm 0.012$ & $12.143 \pm 0.013$ & $11.831 \pm 0.014$ & $11.453 \pm 0.016$ & 29.77h \\
        Ours-sNMC & $6.436 \pm 0.012$ & $12.476 \pm 0.019$ & $13.664 \pm 0.028$ & $13.340 \pm 0.031$ & $12.813 \pm 0.032$ & 29.77h
    \end{tabular}
    \label{combinedspcesnmc1}
    \caption{sPCE and sNMC at $T = 30$ computed using $L = 1e6$, and average agent training time for the location finding experiment. Means and standard errors are from 20000 rollouts, spread evenly across 10 agents trained under unique random seeds. The average training time is across these 10 agents.}
\end{table}

\begin{table}[H]
    \centering
    \begin{tabular}{cccccccc}
         & \multicolumn{2}{c}{} & Time \\
        \hline
        & $\nu = 0.005$ & $\nu = 0.01$ & $\nu = 0.005$ \\
        \hline
        Ours-sPCE & $13.938 \pm 0.021$ & $12.544 \pm 0.022$ & 18.16h \\
        Ours-sNMC & $21.210 \pm 0.165$ & $13.875 \pm 0.049$ & 18.16h
    \end{tabular}
    \label{combinedspcesnmc2}
    \caption{sPCE and sNMC at $T = 10$ computed using $L = 1e7$, and average agent training time for the CES experiment. Means and standard errors are from 20000 rollouts, spread evenly across 10 agents trained under unique random seeds. The average training time is across these 10 agents.}
\end{table}

\section{\raggedright{Experimental Design Problems}}

This appendix explains the experimental design problems in more detail, including their probabilistic models and parameters used in our experiments.

Our code is adapted from the repository by \citet{blau2022optimizing}, which consists of experimental design problems built under the Pyro probabilistic programming language \citep{bingham2018pyro}. All experimental design problems now allow for generalisability testing.

\subsection{Location Finding}

There are $K$ objects on a $d$-dimensional space, and in this experiment we need to identify their locations $\boldsymbol{\theta} = \{\boldsymbol{\beta}_{i}\}_{i = 1}^{K}$ based on the signals that the objects emit. We select designs $\xi$, which are the coordinates chosen to observe the signal intensity, in an effort to learn the locations of the objects. Our spaces are restricted in $\xi \in [-4, 4]^{d}$ to make the problem more tractable. $T = 30$ experiments are conducted both during training and during deployment time, which means that agents can only afford to perform 30 experiments to achieve the highest information gain possible.

The total intensity at point $\xi$ is the superposition of the individual intensities for each object, $$\mu(\boldsymbol{\theta}, \xi) = b + \sum_{i = 1}^{K} \frac{\alpha}{m + ||\boldsymbol{\beta}_{i} - \xi||^{2}},$$ where $\alpha$ is a constant, $b > 0$ is a constant controlling the background signal, and $m > 0$ is a constant controlling the maximum signal. The total intensity is then used in the likelihood function calculation.

For an object $\boldsymbol{\beta}_{i} \in \mathbb{R}^{d}$, we use a standard normal prior given by $$\boldsymbol{\beta}_{i} \sim \mathcal{N}_{d}(\boldsymbol{0}, I),$$ where $\boldsymbol{0}$ is the mean vector, and $I$ is the covariance matrix, an identity matrix, both with dimension $d$.

The likelihood function is the logarithm of the total signal intensity $\mu(\boldsymbol{\theta}, \xi)$ with Gaussian noise $\sigma$. For a given design $\xi$, the likelihood function is given by $$y \mid \boldsymbol{\theta}, \xi \sim \mathcal{N}(\log \mu(\boldsymbol{\theta}, \xi), \sigma^2).$$

The parameters we used are provided in the table below, and we note that $K$ differs at deployment time for our generalisability tests:

\begin{table}[H]
    \centering
    \begin{tabular}{cc}
    \hline
    Parameter & Value \\
    \hline
    $K$ & 2 \\
    $d$ & 2 \\
    $\alpha$ & 1 \\
    $b$ & 0.1 \\
    $m$ & 0.0001 \\
    $\sigma$ & 0.5
    \end{tabular}
    \caption{Location Finding Parameters}
\end{table}

\subsection{Constant Elasticity of Substitution}

We have two baskets $\boldsymbol{x}, \boldsymbol{x'} \in [0, 100]^{3}$ of goods, and a human indicates their preference of the two baskets on a sliding 0-1 scale. The CES model \citep{arrowchen} with latent variables $(\rho, \boldsymbol{\alpha}, u)$, which all characterise the human’s utility or preferences for the different items, is then used to measure the difference in utility of the baskets. We select the two baskets $\boldsymbol{x}, \boldsymbol{x'}$ in a way that allows us to infer the human's preferences. $T = 10$ experiments are conducted both during training and deployment time.

The CES model \citep{arrowchen} defines the utility $U(\boldsymbol{x})$ for a basket of goods $\boldsymbol{x}$ as, $$U(\boldsymbol{x}) = \left( \sum_{i} x_i^{\rho} \alpha_{i} \right)^{\frac{1}{\rho}},$$ where $\rho$ and $\boldsymbol{\alpha}$ are latent variables defined with the prior distributions explained below. This utility function, which is a measure of satisfaction in economic terms, is then used in the likelihood function calculation.

To formulate our CES experiment in the Bayesian framework, we need to define a prior distribution for each of the latent variables. We follow the lead of \citet{foster2020unified} and \citet{blau2022optimizing} by using the following priors for $(\rho, \boldsymbol{\alpha}, u)$,
\begin{align*}
\rho & \sim \text{Beta}(1, 1) \\
\boldsymbol{\alpha} & \sim \text{Dirichlet}([1, 1, 1]) \\
u & \sim \text{Log-Normal}(1, 3^{2}).
\end{align*}
The likelihood function is the preference of the human on a sliding 0-1 scale, which is based on $U(\boldsymbol{x}) - U(\boldsymbol{x'})$. For a given design $\xi$, the likelihood function is given by,
\begin{align*}
\mu_{\eta} & = u \cdot (U(\boldsymbol{x}) - U(\boldsymbol{x'})) \\
\sigma_{\eta} & = \nu u \cdot (1 + \|\boldsymbol{x} - \boldsymbol{x'}\|) \\
\eta & \sim \mathcal{N}(\mu_{\eta}, \sigma_{\eta}^{2}) \\
y & = \text{clip}(s(\eta), \epsilon, 1 - \epsilon),
\end{align*}
where $\nu = 0.005$, $\epsilon = 2^{-22}$, and $s(x) = \frac{1}{1 + e^{-x}}$ is the sigmoid function. Notice that the normally distributed $\eta$ is passed through the sigmoid function, bounding $\eta \in [0, 1]$. Censoring/clipping is applied, which limits the distribution by setting values above $u = 1 - \epsilon$ to be equal to $1 - \epsilon$, and values below $l = \epsilon$ to be equal to $\epsilon$.

The only parameter that changes for our generalisability tests is $\nu$, where we test for $\nu = 0.01$ too. Other parameters remain the same during training and deployment.



\section{\raggedright{Training Performance Plots}}

The plots that follow are the training performance of each agent under the hyperparameter combinations explained in Appendix C. They all explain the rewards, or sPCE, that the agents are able to achieve (on average) at a particular point in training. These are averaged across 10 agents trained under unique random seeds, as done in all of the statistics presented in this paper.

\subsection{Location Finding}

\begin{figure}[H]
\centering
\begin{tabular}{cc}
  \includegraphics[width=75mm]{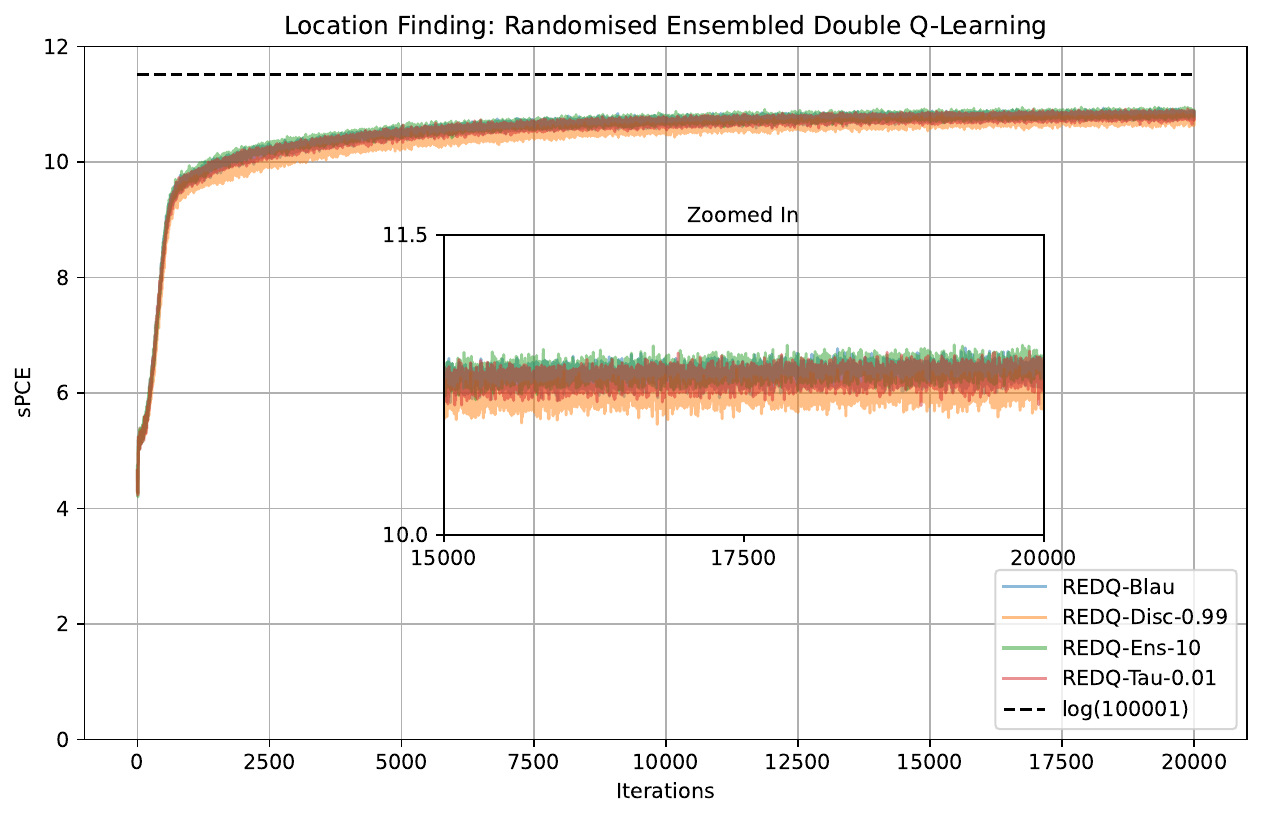} &   \includegraphics[width=75mm]{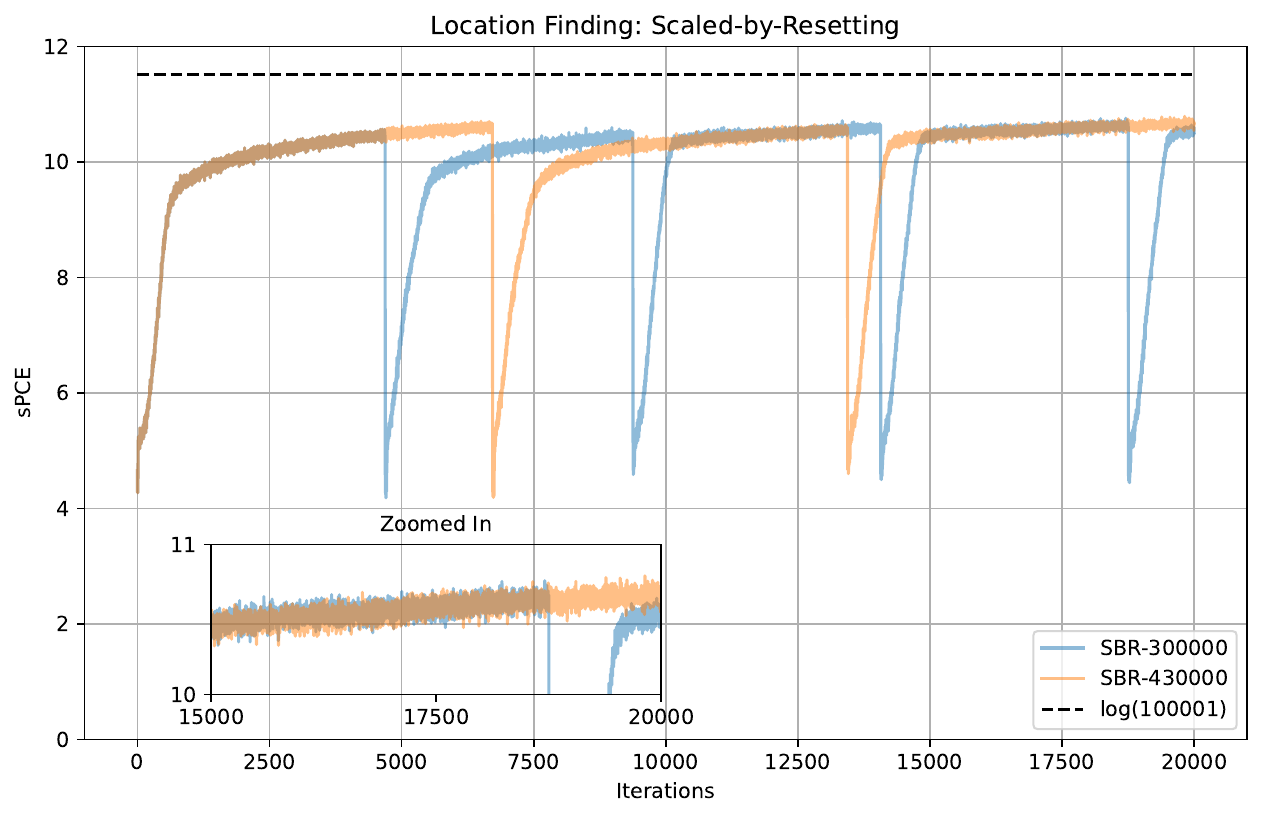} \\
(a) REDQ & (b) DroQ \\[6pt]
  \includegraphics[width=75mm]{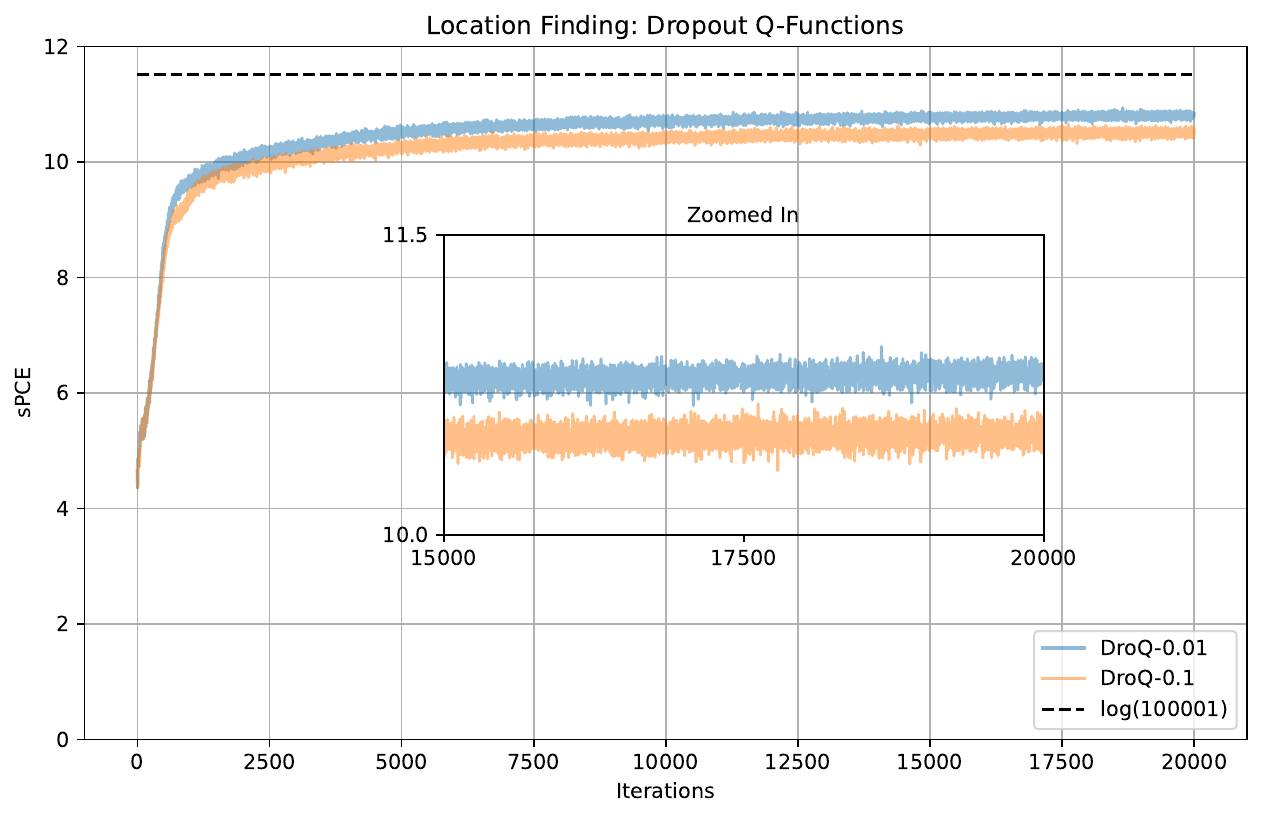} &   \includegraphics[width=75mm]{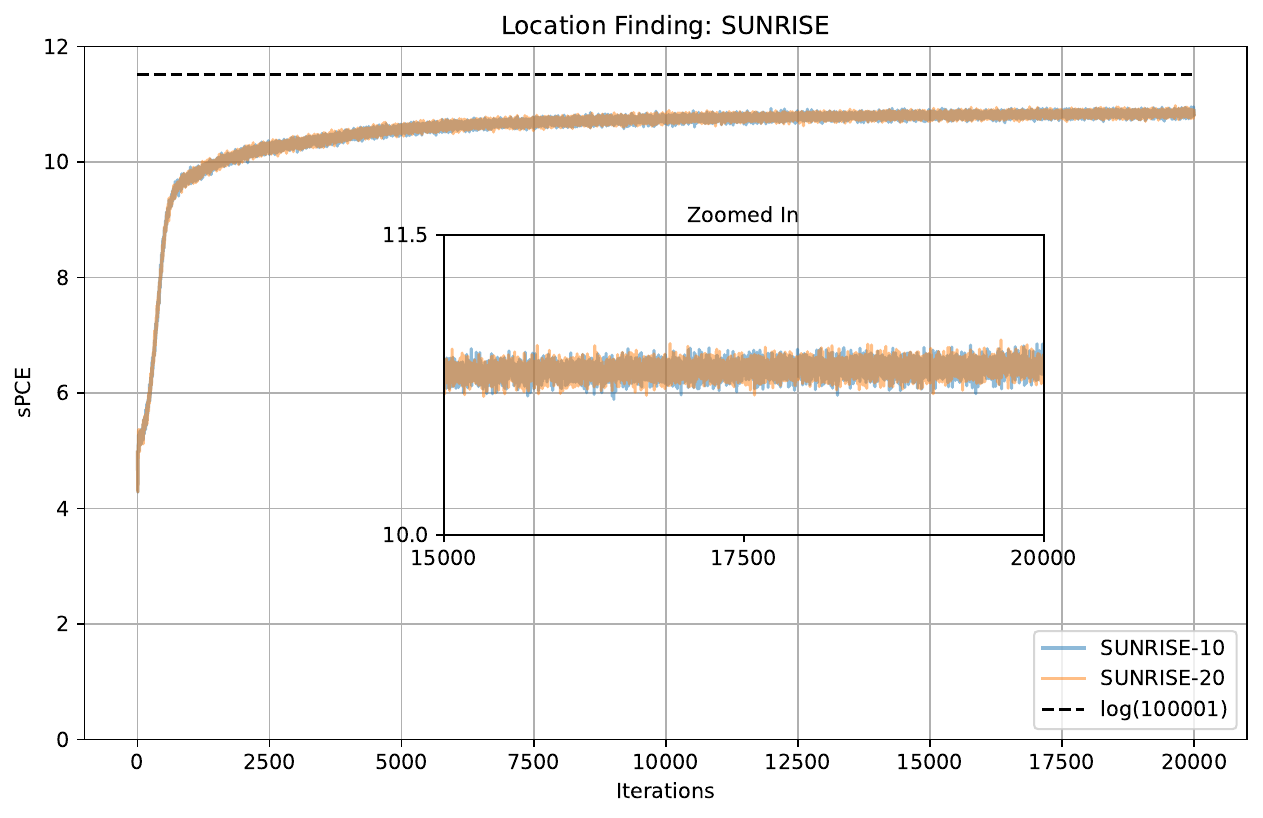} \\
(c) SBR & (d) SUNRISE \\[6pt]
\end{tabular}
\includegraphics[width=75mm]{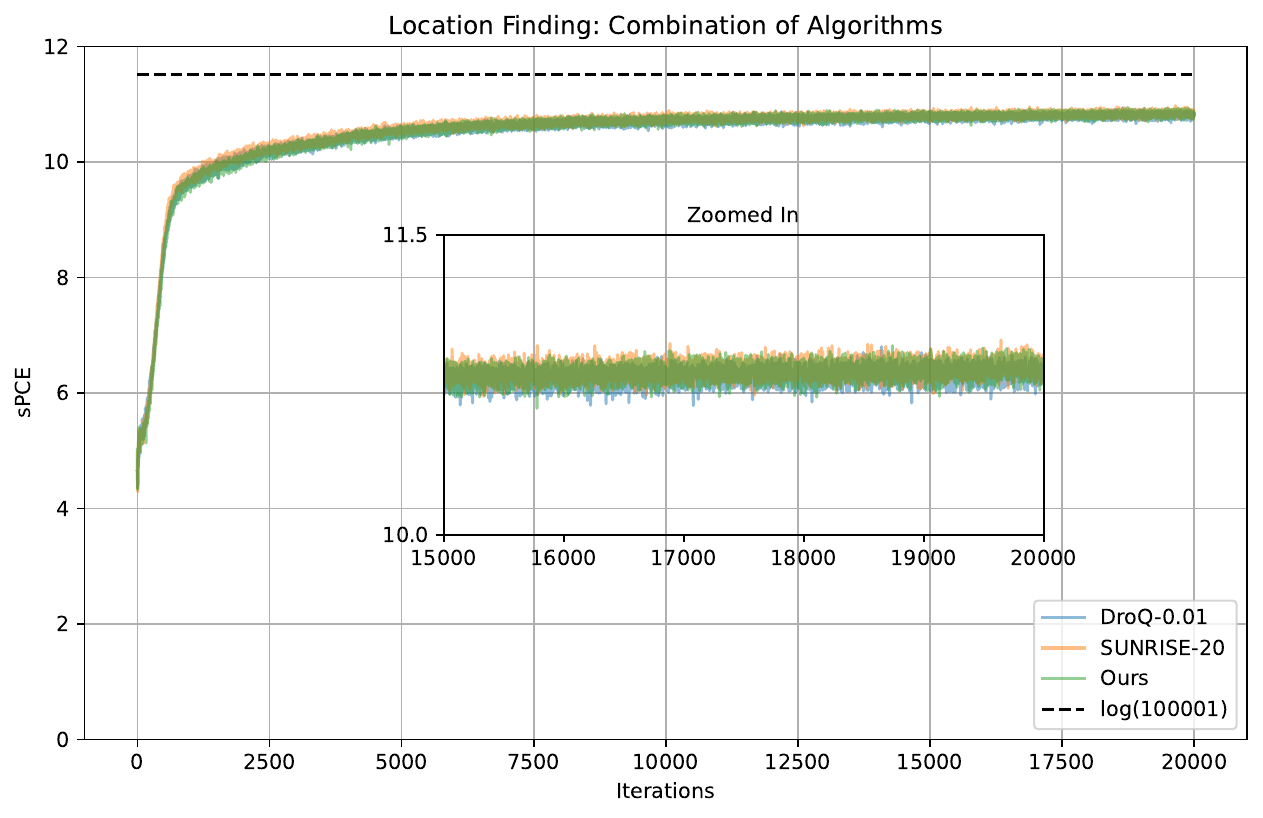} \\
(e) SUNRISE-DroQ
\caption{sPCE at $T = 30$ training performance, under several hyperparameter combinations and algorithms, for the location finding experiment. sPCE is computed using $L = 1e5$, displayed as a dotted line on the plots.}
\end{figure}


\subsection{Constant Elasticity of Substitution}

\begin{figure}[H]
\centering
\begin{tabular}{cc}
  \includegraphics[width=75mm]{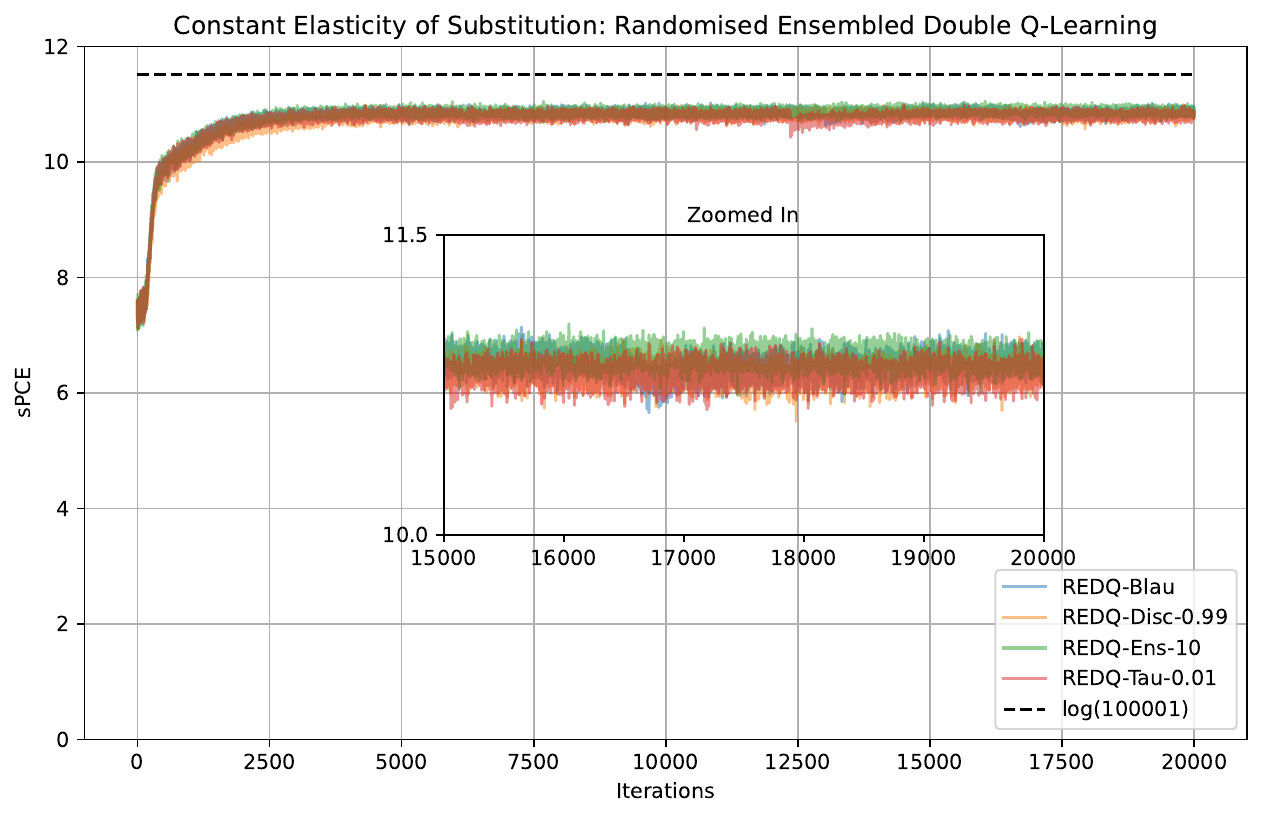} &   \includegraphics[width=75mm]{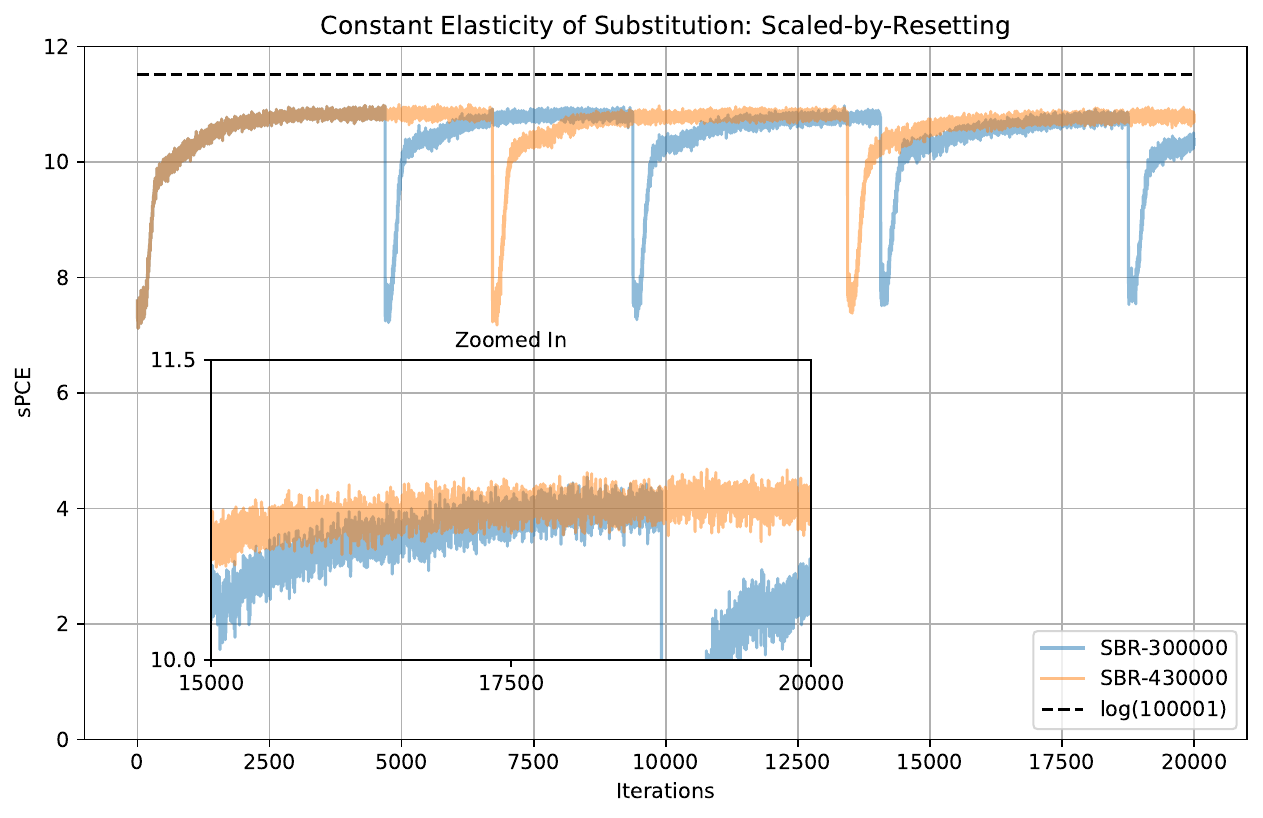} \\
(a) REDQ & (b) DroQ \\[6pt]
  \includegraphics[width=75mm]{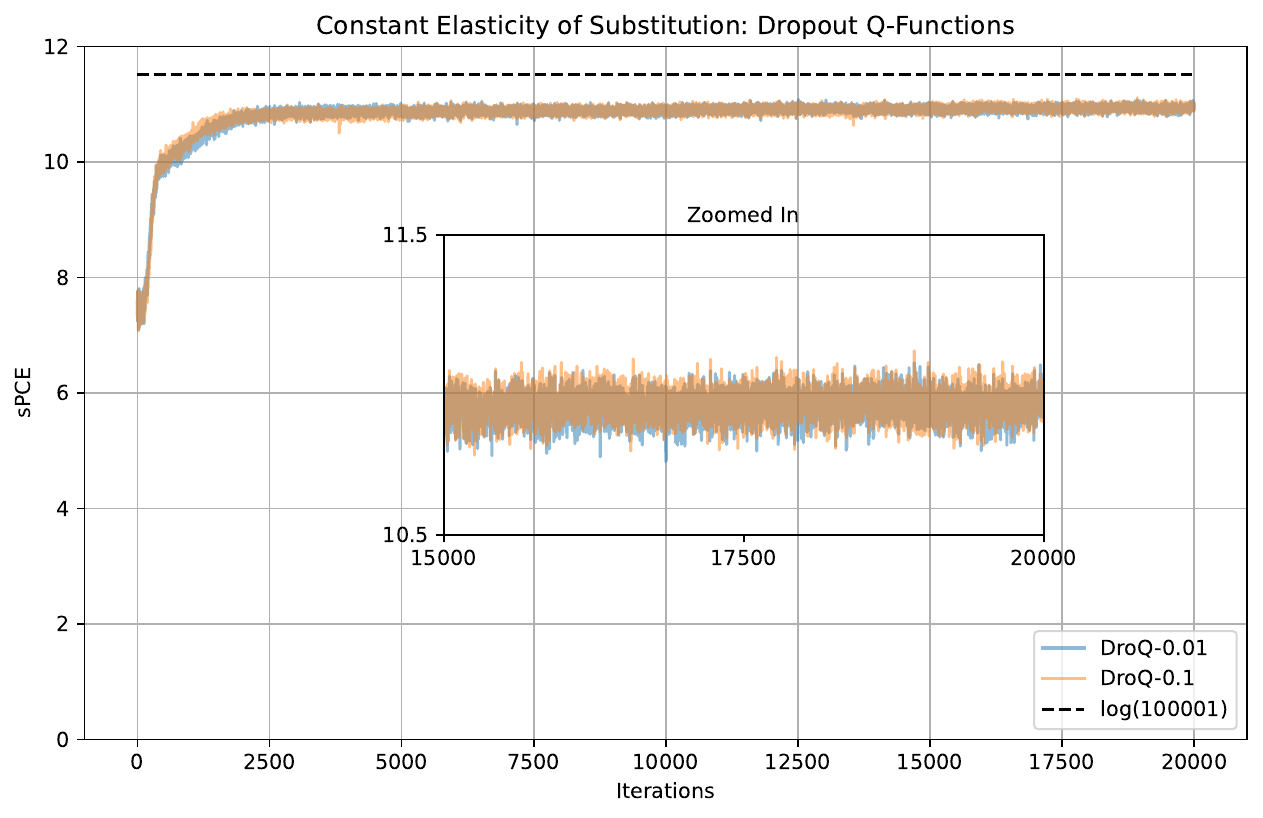} &   \includegraphics[width=75mm]{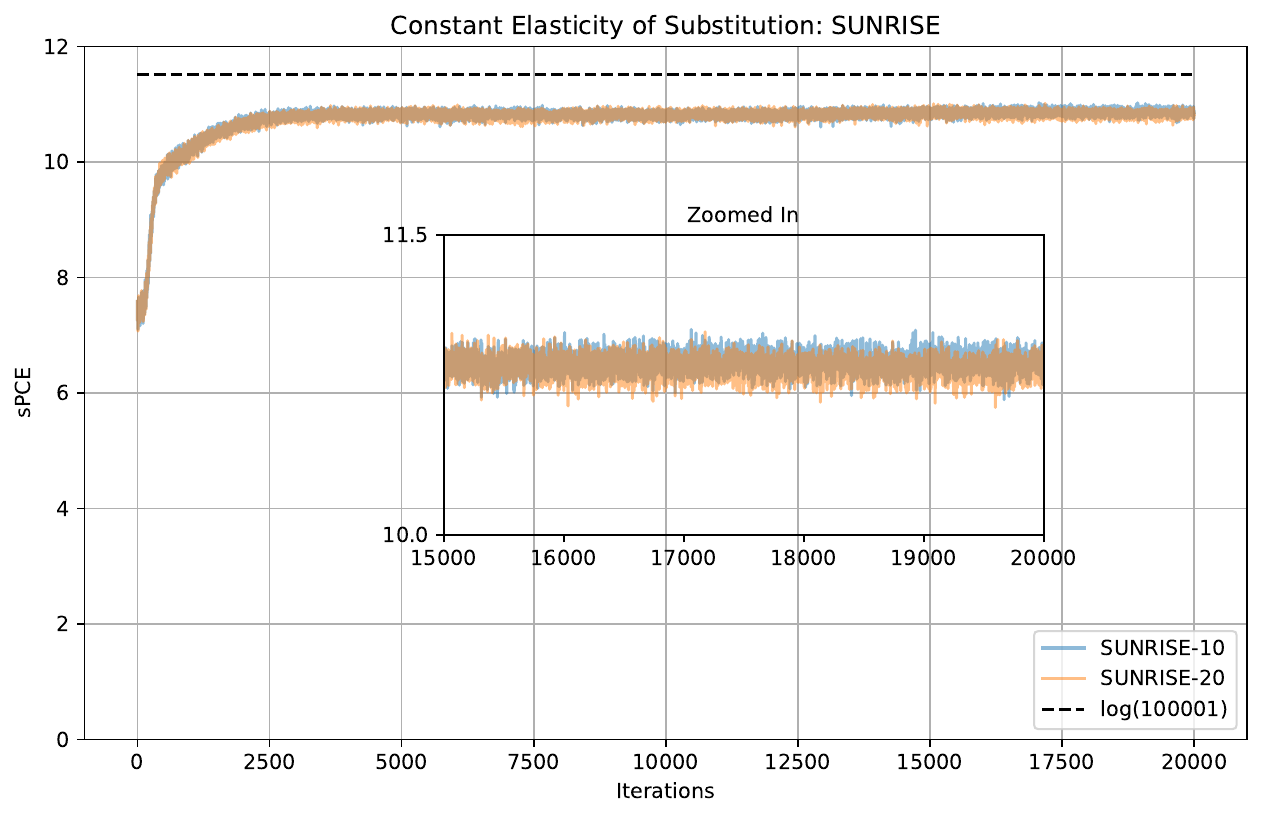} \\
(c) SBR & (d) SUNRISE \\[6pt]
\end{tabular}
\includegraphics[width=75mm]{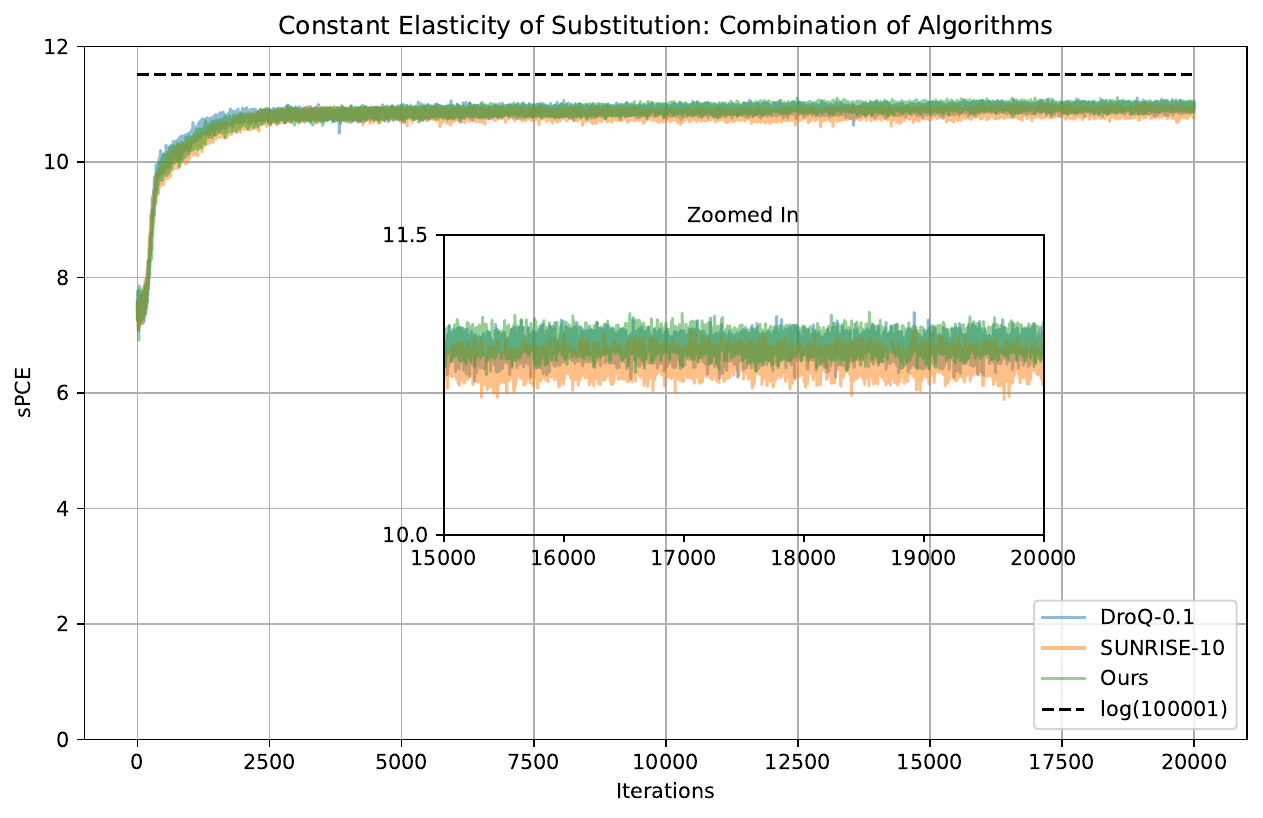} \\
(e) SUNRISE-DroQ
\caption{sPCE at $T = 10$ training performance, under several hyperparameter combinations and algorithms, for the CES experiment. sPCE is computed using $L = 1e5$, displayed as a dotted line on the plots.}
\end{figure}

\section{\raggedright{SUNRISE Tweaks}}


We explain how we tweaked the SUNRISE algorithm at deployment time, in an effort to offer more informative design choices.

One major change from our SUNRISE implementation and that explored by \citet{lee2021sunrise} is how we use our ensemble of agents at deployment time. \citet{lee2021sunrise} choose to average the means of the TanhNormal distributions modelled by each ensemble policy, and use this average as the next design to select. We empirically find that performance severely falls behind that of the other algorithms through this evaluation method, at least for the location finding experiment. We therefore propose two alternatives to evaluating the agents: 

\begin{enumerate}
    \item Randomly selecting a policy to sample an action from;
    \item Forming a new TanhNormal distribution to sample actions from based on the average means and average variances of the ensemble policies. The average of the means is used as the mean, and the average of the variances is used as the variance for the new TanhNormal distribution.
\end{enumerate}

In the results tables displayed for each experimental design problem, SUNRISE-20-A is the agent that follows the evaluation method by \citet{lee2021sunrise} using $\delta = 20$. A similar naming convention is used for the other evaluation methods of SUNRISE-20. SUNRISE-20-B are two distinct agents that randomly select a policy to sample an action from at deployment time, our first alternative method. SUNRISE-20-C uses the average means and average variances to formulate a new policy to sample actions from, our second alternative method. The underlying trained agent here is SUNRISE-20. The only change is how we deploy SUNRISE-20 at deployment time through the methods explained.

\subsection{Location Finding}

We find that the first evaluation method yields superior performance for both sPCE and sNMC, since we are sampling an action from a real distribution learnt during training, and not one that was combined through simply constructing a new distribution based on averaging. Of course, other methods could be explored instead of averaging, or by forming an alternative distribution. The results presented in the main body of this paper, and before this appendix, follow the first evaluation method we propose.

SUNRISE-20-A performs the worst across the board, as we find empirically. We can safely conclude that, at least in our investigated approach, that the evaluation method proposed by \citet{lee2021sunrise} does not offer the impressive results that SUNRISE seeks to deliver. SUNRISE-20-C, while it performs slightly better, also fails to achieve useful values of sPCE. SUNRISE-20-B offers performance comparable to the other algorithms investigated.

\begin{table}[H]
    \centering
    \begin{tabular}{cccccccc}
         & \multicolumn{5}{c}{sPCE} \\
        \hline
        & $K = 1$ & $K = 2$ & $K = 3$ & $K = 4$ & $K = 5$ \\
        \hline
        SUNRISE-20-A & $5.947 \pm 0.011$ & $8.761 \pm 0.019$ & $8.533 \pm 0.020$ & $7.972 \pm 0.020$ & $7.539 \pm 0.020$ \\
        SUNRISE-20-B & \textbf{6.340 $\pm$ 0.013} & \textbf{11.837 $\pm$ 0.012} & \textbf{12.133 $\pm$ 0.013} & \textbf{11.846 $\pm$ 0.014} & \textbf{11.445 $\pm$ 0.016} \\
        SUNRISE-20-C & $6.125 \pm 0.010$ & $9.563 \pm 0.019$ & $9.597 \pm 0.020$ & $9.393 \pm 0.021$ & $9.110 \pm 0.021$
    \end{tabular}
    \caption{sPCE results for different SUNRISE evaluation methods at $T = 30$ computed using $L = 1e6$, for the location finding experiment. Means and standard errors are from 20000 rollouts, spread evenly across 10 agents trained under unique random seeds.}
\end{table}

\begin{table}[H]
    \centering
    \begin{tabular}{cccccccc}
         & \multicolumn{5}{c}{sNMC} \\
        \hline
        & $K = 1$ & $K = 2$ & $K = 3$ & $K = 4$ & $K = 5$ \\
        \hline
        SUNRISE-20-A & $5.949 \pm 0.011$ & $8.896 \pm 0.021$ & $8.695 \pm 0.023$ & $8.074 \pm 0.022$ & $7.625 \pm 0.021$ \\
        SUNRISE-20-B & \textbf{6.342 $\pm$ 0.013} & \textbf{12.555 $\pm$ 0.019} & \textbf{13.656 $\pm$ 0.029} & \textbf{13.345 $\pm$ 0.030} & \textbf{12.823 $\pm$ 0.032} \\
        SUNRISE-20-C & $6.126 \pm 0.010$ & $9.792 \pm 0.022$ & $10.070 \pm 0.027$ & $9.919 \pm 0.029$ & $9.600 \pm 0.029$
    \end{tabular}
    \caption{sNMC results for different SUNRISE evaluation methods at $T = 30$ computed using $L = 1e6$, for the location finding experiment. Means and standard errors are from 20000 rollouts, spread evenly across 10 agents trained under unique random seeds.}
\end{table}

\subsection{Constant Elasticity of Substitution}

For the CES experiment, the first method performs slightly worse than the evaluation method by \citet{lee2021sunrise}, at least in terms of sPCE. sNMC is largest for our first evaluation method, suggesting that the true EIG falls within a larger range between the bounds -- and can potentially be greater than the true EIG under the other two methods. The sNMC standard errors are the largest for our evaluation method, though not significantly.

Whilst our second proposed method is consistent between both the location finding and CES experiments, the other two are not in terms of sPCE. One may argue in favour of our first method, as we do, since we achieve the best sPCE and sNMC performance on the location finding experiment, and only fail to achieve the highest sPCE for the CES experiment. To stay consistent, we stick to our first evaluation method across both experimental design problems tackled in this paper, unless otherwise stated.

\begin{table}[H]
    \centering
    \begin{tabular}{cccccccc}
         & \multicolumn{2}{c}{sPCE} & \multicolumn{2}{c}{sNMC} \\
        \hline
        & $\nu = 0.005$ & $\nu = 0.01$ & $\nu = 0.005$ & $\nu = 0.01$ \\
        \hline
        SUNRISE-20-A & \textbf{13.806 $\pm$ 0.022} & \textbf{12.277 $\pm$ 0.022} & $17.715 \pm 0.092$ & $12.851 \pm 0.032$ \\
        SUNRISE-20-B & $13.633 \pm 0.022$ & $12.261 \pm 0.023$ & \textbf{19.815 $\pm$ 0.130} & \textbf{13.344 $\pm$ 0.042} \\
        SUNRISE-20-C & $11.252 \pm 0.035$ & $10.062 \pm 0.032$ & $13.816 \pm 0.066$ & $10.418 \pm 0.036$
    \end{tabular}
    \caption{sPCE and sNMC results for different SUNRISE evaluation methods at $T = 10$ computed using $L = 1e7$, for the CES experiment. Means and standard errors are from 20000 rollouts, spread evenly across 10 agents trained under unique random seeds.}
\end{table}


\section{\raggedright{Source Code and Hardware}}

In this appendix, we explain how to find our source code, and the hardware we used in our experiments.

\subsection{Source Code}

Our code is publicly available as a GitHub repository at https://github.com/yasirbarlas/RL-BOED. As done by \citet{blau2022optimizing}, we utilise Pyro \citep{bingham2018pyro}, Garage \citep{garage}, and PyTorch \citep{paszke2019pytorch}. Our repository is built over that by \citet{blau2022optimizing}, which can be found at \url{https://github.com/csiro-mlai/RL-BOED}. Any additions and changes are noted in the \textit{README.md} file of our repository.

\subsection{Hardware}

All experiments were run on the Hyperion High-Performance Computer at City St George's, University of London. A single NVIDIA A100 80GB PCIe GPU or NVIDIA A100 40GB PCIe GPU (only the VRAM differs) was used for each experiment, through the SLURM Workload Manager. 4 CPU cores and 40GB of RAM were assigned to each agent for training. An Intel(R) Xeon(R) Gold 6248R CPU @ 3.00GHz was used.

\end{document}